\documentclass{article} % For LaTeX2e
\usepackage{iclr2024_conference,times}

\usepackage[hidelinks]{hyperref}
\usepackage{url}

\usepackage{booktabs}
\usepackage{graphicx}
\usepackage{subcaption}
\usepackage{enumitem}
\usepackage{xcolor}
\usepackage{colortbl}
\usepackage{amsmath}
\usepackage{amssymb}
\usepackage{mathabx,bm}
\usepackage{tabularx}

\title{RewriteLM: An Instruction-Tuned Large Language Model for Text Rewriting}
\author{
Lei Shu\thanks{Equal Contribution. Correspondence: \{leishu,leimeng\}@google.com. }\quad
Liangchen Luo\quad
Jayakumar Hoskere\quad
Yun Zhu\quad
Yinxiao Liu\quad
Simon Tong\And
Jindong Chen\quad
Lei Meng\footnotemark[1]\\\\
Google Research
}

\newcommand{\eg}{\textit{e}.\textit{g}.}
\newcommand{\ie}{\textit{i}.\textit{e}.}

\iclrfinalcopy % Uncomment for camera-ready version, but NOT for submission.
\begin{document}

\maketitle

% Main body.

% Author contributions and acknowledgements.


\begin{abstract}
Large Language Models (LLMs) have demonstrated impressive  capabilities in creative tasks such as storytelling and E-mail generation. However, as LLMs are primarily trained on final text results rather than intermediate revisions, it might be challenging for them to perform text rewriting tasks. Most studies in the rewriting tasks focus on a particular transformation type within the boundaries of single sentences. In this work, we develop new strategies for instruction tuning and reinforcement learning to better align LLMs for cross-sentence rewriting tasks using diverse wording and structures expressed through natural languages including 1) generating rewriting instruction data from Wiki edits and public corpus through instruction generation and chain-of-thought prompting; 2) collecting comparison data for reward model training through a new ranking function. To facilitate this research, we introduce \textsc{OpenRewriteEval}, a novel benchmark covers a wide variety of rewriting types expressed through natural language instructions. Our results show significant improvements over a variety of baselines. The public repository is available on GitHub under Google Research\footnote{This paper is accepted by AAAI 2024. \url{https://github.com/google-research/google-research/tree/master/rewritelm}.}.
\end{abstract}

\section{Introduction}
\label{sec:intro}
\begin{table*}[h]
\small
\centering
\begin{tabular}{p{12cm}}
\toprule
    \textbf{Source Text:}  The way in which we feed our children determines how healthy they are. It also determines how nicely they develop and develop. Children need to be fed a selection of foods each day in order to make sure their physical, emotional, and mental health. All foods are from 1 of the basic food groups. Milk and cheese come from the dairy group, for instance, and green beans come from the vegetable group. Bread comes from the grain group and beef comes from the meat group. Chocolate comes from the fat and sugars group. Our children need so many servings per day from each of these food groups to maintain good health. \\
    \midrule
    \textbf{Instruction:} Rewrite the text so that it is easy to understand. \\
    \midrule
    \textbf{Target Text:}  What we give our kids to eat affects how healthy they are and how they grow. Kids need to eat different types of food each day to stay physically, emotionally, and mentally healthy. All foods belong to one of the five food groups: dairy, vegetables, grains, meat and beans, and fruits. Milk and cheese are dairy foods, green beans are vegetables, bread is a grain, beef is a meat, and chocolate is a fat and sugar food. To stay healthy, kids need to eat a certain number of servings from each food group every day. \\
    \midrule
    \textbf{Instruction:} Summarize the text.  \\
    \midrule
    \textbf{Target Text:}  Feeding children a variety of foods from the five fundamental food groups (dairy, vegetables, grains, meat, and fats/sweets) is essential for their overall health and development, including their physical, emotional, and mental well-being. \\
    \bottomrule
    \end{tabular}
    \caption{Rewrite examples. The input for RewriteLM contains an instruction and a source text. The target is a rewrite. We present one source text and two instructions here for illustration purpose.}
    \label{tab:rewriteexample}
\end{table*}
% \begin{figure*}
% \includegraphics[width=0.8\textwidth]{figures/rewrite_example.pdf}
% \centering
% \caption{Rewrite examples. The input for RewriteLM contains an instruction and a source text. The target is a rewrite. We present one source text and two instructions here for illustration purpose.}
% \label{fig:rewriteexample}
% \end{figure*}

Text rewriting plays an essential role in a wide range of professional and personal written communications. It can be conceptualized as a form of controllable text generation \citep{zhang2022survey}, %\yz{citation?}
where a specified textual input is modified based on the user's requirement. Several categories of text rewriting have been extensively researched, such as paraphrasing \citep{siddique2020unsupervised, xu2012paraphrasing},  style transfer \citep{
riley2020textsettr, 
zhang2020parallel,
reif2021recipe}, and sentence fusion \citep{mallinson2022edit5}.

Recent advances in Large Language Models (LLMs) have shown impressive zero-shot capabilities in a wide range of text generation tasks expressed through natural language instructions  \citep{chung2022scaling}. However, user expectation for text rewriting is high and any unintended edits by the model negatively impact the user's satisfaction. Given that the LLMs can be hard to control \citep{qin2023chatgpt} and prone to generating ``hallucinated'' content \citep{ji2023survey}, we propose new methods
to ensure that the model is properly 
trained and evaluated. %using instruction datasets
%\jd{How do we show the importance of using instruction dataset? Can we point to any reference?}
%that are both diverse and representative.

%To address this, 

We present a strong model --- RewriteLM, an instruction-tuned large language model for cross-sentence text rewriting. Similar to InstructGPT \citep{ouyang2022training}, we perform supervised fine-tuning and reinforcement learning (RL) steps. 
% (1) fine-tuning on Wiki-based instruction dataset and synthetic instruction dataset; 
% (2) training the reward model on human and synthetic preference data; and 
% (3) optimizing a policy by training it against a reward model using reinforcement learning.
However, InstructGPT heavily relies on human labelers to collect instruction data (for supervised fine-tuning) and comparison data (for reward training). We instead develop the following data and training strategies with minimal human intervention: 
1) For instruction tuning, we extract cross-sentence edits from Wiki edits and improves its instruction. To improve the diversity of the dataset, we also generate synthetic instruction and rewrite output from public corpus through chain-of-thought prompting and post-processing 2) For reinforcement learning, we collect comparison data for reward model training through a new ranking function, which assess the quality of rewrite along several dimensions including content preservation, hallucination, linguistic variability, and length text change.

To properly test the capability of RewriteLM, we introduce a new benchmark \textsc{OpenRewriteEval} by collecting human-generated text rewrites with natural language instructions. Unlike the previous benchmarks for text rewriting, which mostly had restricted types \citep{reif2021recipe, mallinson2022edit5} and performed within the boundaries of single sentences \citep{riley2020textsettr, siddique2020unsupervised, mallinson2022edit5}, our benchmark is designed for research on cross-sentence text rewrite and covers a wide variety of rewriting types expressed through natural language instructions.

% (1) For instruction data generation, we extract cross-sentence, high quality edits with substantial changes from Wiki edits as our main natural source \citep{schick2022peer} of the source and target pairs, and use their associated comments as proxies for instructions.
% (2) To increase the amount and the diversity of the instruction data beyond Wikipedia, we leverage Chain-of-thoughts (CoT) and the capability of the LLMs to generate instructions and target text for various input text.
% (3) To generate more preference data, we sample multiple LLM model outputs and rank them using a human-designed heuristic ranker.
% We then fine-tune the pretrained language models on the collected data to produce RewriteLM models.

We conduct empirical studies to evaluate the model performance on the \textsc{OpenRewriteEval} benchmark.
The results show that even current state-of-the-art pretrained LLMs have poor performance on open-ended rewriting tasks.
LLMs fine-tuned on general-purpose instruction datasets like Flan-PaLM \citep{chung2022scaling} and Alpaca \citep{alpaca} have better performance compared with the pretrained foundation models, but still have room for improvement.
The proposed RewriteLMs, including Rewrite-PaLM and Rewrite-PaLM 2, both outperform their corresponding foundation models by a significant margin.
They also outperform other instruction-tuned LLMs, showcasing the effectiveness of the generated training data.
Applying reinforcement learning on top of the supervised tuned Rewrite-PaLM 2 further improves its performance, resulting in a new state-of-the-art model Rewrite-RL$_\text{r/w}$-PaLM 2 for text rewriting.

%Our results demonstrate that our model is good at keeping all the essential content and meaning of the source text, while also able to rewrite with varied language and structures.

%Unlike InstructGPT that heavily replies on human labelers to collect instruction data (for supervised fine-tuning) and preference data (for reward training), we develop the following strategies to minimize human intervention in the data collection process: 1) For instruction data generation, we extract long-form, high quality edits with substantial changes from Wiki as our main natural source \citep{schick2022peer} of training data  and its associated comments (refined through LLMs) as a proxy for instruction; 2) To increase the amount and the diversity of the instruction data beyond Wiki, we use self-traing \yz{citation?} technique to utilize the capabilities of LLMs. This involves generating instructions and target texts using LLMs and pruning. 3) To generate preference data, we sample multiple model outputs and rank those outputs using human-designed heuristic ranker.

Our main contributions can be summarized as follows:
\begin{itemize}
    \item A new benchmark, \textsc{OpenRewriteEval}, designed for research on cross-sentence rewrite  and covering a wide variety of rewriting types expressed through natural language instructions, such as formality, expansion, conciseness, paraphrasing, tone and style transfer. Unlike previous benchmarks, which were primarily focused on specific rewrite types within the boundaries of single sentences, our benchmark is specifically designed to facilitate cross-sentence rewrites with open-ended natural language instructions. To the best of our knowledge, no such dataset has existed previously.    
    \item New strategies for instruction tuning and reinforcement learning to better align LLMs for cross-sentence rewriting tasks using diverse wording and structures expressed through natural languages including 1) generating rewriting instruction data from Wiki edits and public corpus through instruction generation and chain-of-thought prompting 2) collecting comparison data for reward model training through a new ranking function. We demonstrate that RewriteLM model achieved the state-of-the-art performance in cross-sentence rewriting tasks on OpenRewriteEval. % generating rewriting instruction data from Wikiedits and public corpus through instruction generation andchain-of-thought prompting 2) collecting comparison data forreward model training through a new ranking function.

    %A strong baseline model, RewriteLM, has demonstrated superior performance compared to the state-of-the-art LLMs for text rewriting, especially in long-form. We developed new strategies to generate diverse, long-form rewriting instruction datasets and preference data that can be used to enhance the editing and rewriting capabilities of LLMs. Our results demonstrate that our model has strong performance in rewriting with diverse language and structures while preserving the essential meaning of the original input. Importantly, our model effectively mitigates the issue of ``hallucinations'', which is commonly observed in existing LLMs.
\end{itemize}

\section{Related Work}
\label{sec:related}
% Writing assistant is a popular application that helps real users write in their daily lives. It covers many editing cases like grammar error correction (Grammarly\footnote{\url{https://www.grammarly.com/}}), Etc. Unfortunately, most existing writing assistant application does not support open instruction. Instead, they focus on small-set editing tasks. Chatgpt\footnote{\url{https://openai.com/blog/chatgpt}} is an open-ended content generation tool. It allows users to insert textual instruction. However, Chatgpt puts more effort into content generation instead of rewriting.
% Cocomposer is one writing assistant that can rewrite the source text based on users' open instructions.

\textbf{Text Editing}. The majority of the research on rewriting currently focuses on a particular set of editing tasks at the sentence level, such as paraphrase~\citep{huggingface:dataset:stsb_multi_mt}, style transfer~\citep{tikhonov2019style}, spelling correction~\citep{napoles2017gec}, formalization~\citep{rao2018dear}, simplification~\citep{xu2016optimizingsari} and elaboration~\citep{iv2022fruit}. 
\cite{faltings2020text} trained an editing model to follow instructions using Wikipedia data. However, their focus was solely on edits limited to a single sentence. PEER~\citep{schick2022peer} can follow human-written instructions for updating text in any domain, but is still limited by the edit types available on Wikipedia. Moreover, it was only evaluated on a small set of edit types from a human-defined instruction evaluation benchmark \citep{dwivedi-edit-2022}.

\textbf{Instruction Tuning}. Instruction tuning has shown to improve model performance and generalization to unseen tasks \citep{chung2022scaling,sanh2022multitask}.
InstructGPT~\citep{ouyang2022training} extends instruction tuning further with reinforcement learning with human feedback (RLHF), which heavily relies on human labelers to collect instruction data and model output rankings for training. The focus of these works was primarily on extensively researched tasks and benchmarks, which do not include open-ended text rewriting.

\textbf{Data Augmentation via LLM}. A common data augmentation approach involves utilizing trained LLMs to generate more data, which is subsequently incorporated as training data to enhance the model's performance \citep{he2019revisiting,xie2020self,huang2022large}. PEER \citep{schick2022peer} leverage LLMs to infill missing data and then use this synthetic data to train other models. Self-Instruct~\citep{wang2022selfinstruct, alpaca} improves its ability to accurately follow instructions by bootstrapping off its own generated outputs. Our work builds upon similar ideas and leverages the power of LLMs to enhance existing datasets and generate additional synthetic datasets.

\section{Methods}
\label{sec:method}

% \begin{figure}
% \includegraphics[width=0.98\textwidth]{figures/ssit_steps.pdf}
% \centering
% \caption{Self-Instruct Learning Steps.}
% \label{fig:step}
% \end{figure}

In this section, we discuss the training data (Section \ref{subsection:training_data}) and the training procedure (Section \ref{training_procedure}) for the proposed RewriteLM models.
Table \ref{tab:training-data-statistic} provides a comprehensive overview of the training data's statistics, while Figure \ref{fig:instruction-plot-training} illustrates the distribution of instructions within the training data.

\subsection{Training Dataset}
\label{subsection:training_data}
\begin{figure*}[h]
\includegraphics[width=0.98\textwidth]{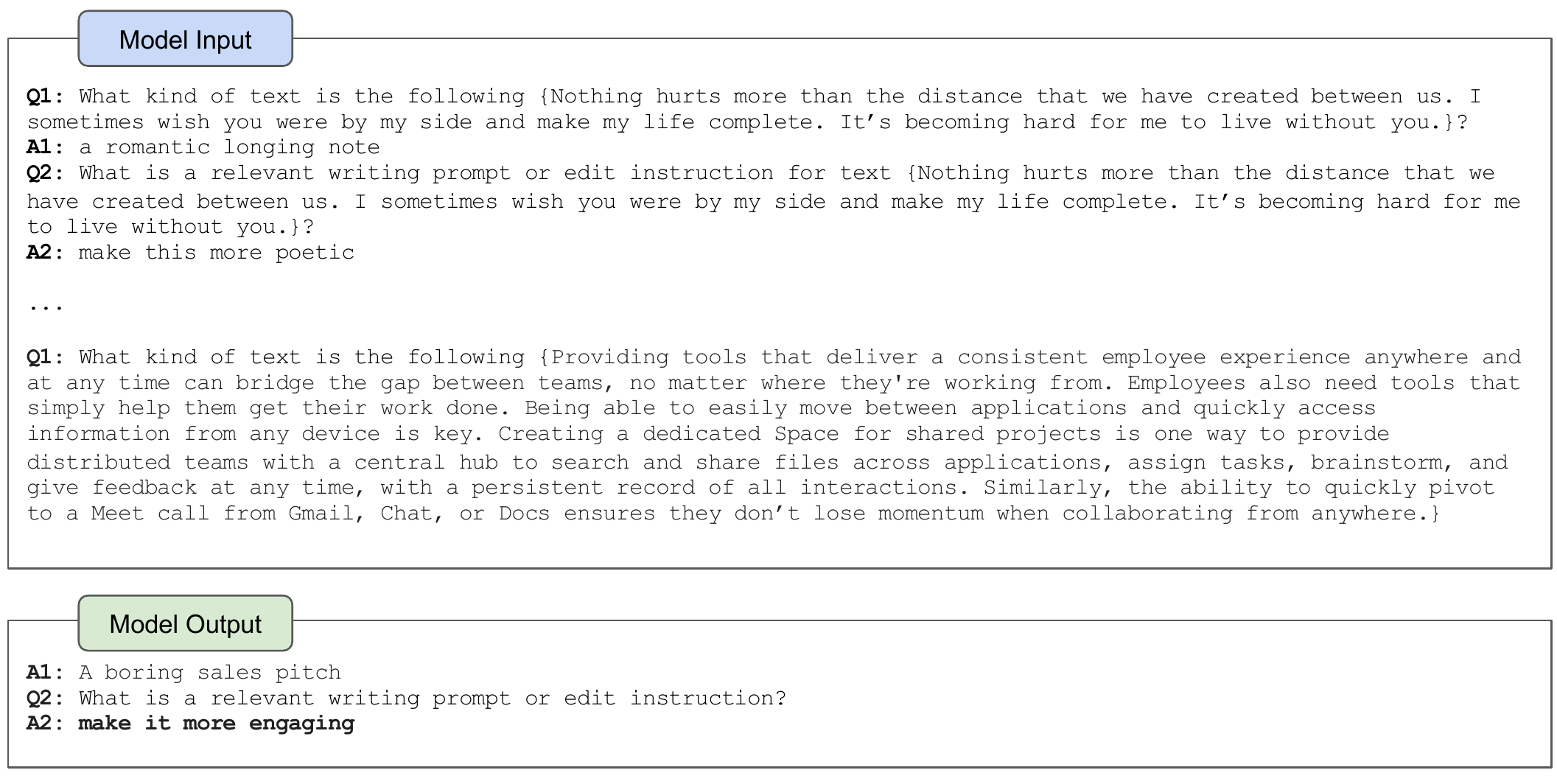}
\centering
\caption{Chain-of-thought (CoT) approach to generating rewrite instructions. The answer to the second question in the output is the generated instruction.}
\label{fig:cot}
\end{figure*}
\subsubsection{Wiki Instruction Dataset}
\label{section:wikiinstructiondata}
We examine Wiki revisions and extract long-form, high quality edits that contain substantial changes. We also use the associated edit summary of the revision as a proxy for the instructions.  We describe edit extraction, edit filtering, and instruction improvement in details:

\begin{itemize}
\item \textbf{Edit Extraction:} We initiate the instruction tuning data collection process by gathering Wikipedia revision history, where each revision record includes the original text, revision differences, and an edit summary written by the revision author. We extract text block differences between each consecutive snapshots of a Wikipedia article and the associated edit summary, following the approach in \citet{schick2022peer}. In the rest of the section, we may use the terms \textit{source text}, \textit{target text} and \textit{comment} to denote the text before revision, the text after revision and the edit summary of a revision record, respectively.

\item \textbf{Edit Filtering:} In order to create long-form, high-quality edits with substantial changes, we remove revision records that meet any of the following criteria: (i) the edit summary indicates low-quality content of a snapshot, such as containing ``revert" or ``vandalism" keywords; (ii) the edit summary contains keywords indicating a format-only change (\eg bold-facing or hyperlinks), which is not a focus of this work; (iii) the source text contains two or fewer sentences. %or (iv) the edit distance ratio (see Section  \ref{subsec:baselines_and_metrics}) between the source and target texts is small. (v) significant new content or 

\item \textbf{Instruction Improvement:} The raw comment may not directly meet our data requirements, which can be empty, contain irrelevant descriptions to the revision, or not describe the editing behavior (\eg only describes the deficiencies of source text). We take the following steps to enhance the quality of the instructions: (i) Extract revision records where the edit summary starts with a verb describing an edit intent (\eg ``make the text easier to read''); (ii) Fine-tune PaLM2-XXS to generate comments from \texttt{<source>-<target>} text pairs as well as learn to control the length and specificity of the instructions. We use the heuristic that if a comment mentions a word from the edit then it is a detailed instruction. (iii) Generate detailed comments for all \texttt{<source>-<target>} pairs using the model trained in the previous steps.
% \yz{I feel the 2/ and 3/ is not easy to understand for readers...Maybe simplify as: 2) Fine-tunes a LLM to learn to generate comment from a source-and-target text pairs. And then generate comments for all source and target pairs using this trained model. 3) Apply the heuristic filtering to get rid of the comments that are too short or have no overlap with the edit.} \lm{We did not apply heuristic to filter comments. They are all generated by LLMs}

\end{itemize}

\subsubsection{Synthetic Instruction Dataset}
\label{section:SyntheticTrainingData}

The Wiki instruction dataset is limited by the available edit types found on Wikipedia. To collect a more diverse and representative instruction dataset, we first use chain-of-thoughts prompting and few-shot prompting to generate instructions, and then generate the target text from a general purpose LLM model:

\begin{table*}[]
\centering
\fontsize{9pt}{9pt}\selectfont
\setlength{\tabcolsep}{1.2pt}
\begin{tabular}{lcccccccc}
\toprule
     & Size    & Inst Len & Src Len  & Tar Len  & Len Ratio & Edit Dist & Edit Ratio & Rouge1  \\\midrule
All  & $24384$ & $6.85$   & $118.86$ & $141.09$ & $1.20$    & $115.44$  & $0.97$     & $60.95$ \\ \midrule
Wiki & $18196$ & $7.38$   & $112.17$ & $98.39$  & $0.90$    & $77.69$   & $0.70$     & $64.77$ \\
Synthetic  & $6188$  & $5.30$   & $138.54$ & $266.63$ & $2.10$    & $226.43$  & $1.78$     & $49.72$ \\
\bottomrule
\end{tabular}
\caption{RewriteLM Training Data Statistics: This table includes statistics for the entire training set (``All"), data derived from Wikipedia (``Wiki", Section~\ref{section:wikiinstructiondata}), and synthetic data generated from large language models (``Synthetic", Section \ref{section:SyntheticTrainingData}). Metrics are the number of examples (Size); the average number of words in instructions (Inst Len), source texts (Src Len), and target texts (Tar Len); the average length fraction (Len Ratio) between the target and source texts; the average edit distance (Edit Dist) between source and target; the ratio of edit distance to source text length (Edit Ratio); and the Rouge1 score comparing source and target texts. All measurements are conducted at the word-level.}
\label{tab:training-data-statistic}
\end{table*}

\begin{itemize}

\item \textbf{Instruction generation: } By applying a 3-shot chain-of-thought (CoT) prompting method to text inputs from any domain (see Figure \ref{fig:cot}), we can leverage the knowledge acquired by the PaLM2-L during pre-training. This enables the LLM to produce more diverse instructions beyond Wiki edit types. CoT contains two QA stages: \textbf{Text description} (answering ``What kind of text is the following'') and \textbf{Instruction generation} (answering ``What is a relevant writing prompt or edit instruction for text'').  The answer to the second question is the generated instruction.

\item \textbf{Target generation:} 
Given the source text and the generated instructions, we generate the model outputs with a general purpose instruction tuned LLM (text-bison-001\footnote{https://cloud.google.com/vertex-ai/docs/generative-ai/model-reference/text}) and filter them in a post-processing step (see Section \ref{subsubsection:Post-processing}). 

%if there are significant hallucinated or missing content  using NLI variants (see Section \ref{subsec:baselines_and_metrics}). If user expect significant change of output length, we could also use length ratio (see Section \ref{subsec:baselines_and_metrics}) to filter.

%\item \textbf{Rewrites filtering:} The language model tends to generate hallucinated content. To avoid this, we use NLI~\citep{bowman2015large} to detect hallucinations. Let $s_1$ and $s_2$ be the NLI scores for 1) using input as the premise and output as the hypotheses and 2) using output as the premise and input as the hypotheses, respectively. For meaning preserved use cases (e.g., rephrase), we require both $s_1$ and $s_2$ to be high. For adding new information use cases, we require $s_2$ to be high and $s_1$ to have a relatively lower score. We expect $s_1$ to be high and $s_2$ to have a relatively lower score for information removal use cases. For data that does not meet the above criteria, we will filter them out. \yz{filtering -> NLI filtering?}
\end{itemize}

\subsubsection{Heuristic Post-Processing}
\label{subsubsection:Post-processing}

In order to improve the quality of the instruction datasets, we do the following post-processing: (1) In general, rewriting should preserve the overall meaning of the text, and thus, we employ Natural Language Inference (NLI; See Section \ref{subsec:baselines_and_metrics}) to detect ``hallucinations'' from the source to the target text and vice versa. If the ``hallucination'' is in the target text and fixable using simple heuristic rules, we remove the ``hallucination" from the target text and keep the instance. (2) For any other detected ``hallucination'', we filter the instance. %\yz{Is it loop detection instead of halluciation detection? should we talk about the detailed n-gram based algorithm here? No.} \lm{This is about removing chattyness} \yz{Chatty is less likely to happen in ULM? maybe we should talk about the loop detection algorithm which seems non-trivial?} \lm{I don't think we remove loop in the training data. It does not matter whether non-trivial or not for the paper. Simpler and novel approach is preferred}
(3) If the difference between the source and target texts is unexpectedly small, we  also filter the instance.

\subsection{Modeling}
\label{training_procedure}

\textbf{Supervised Fine-Tuning (SFT)}. Given a pretrained language model $M_\text{base}$, we fine-tune it using the instruction tuning dataset discussed in Section \ref{subsection:training_data}, producing a model $M_\text{SFT}$. We employ the decoder-only Transformer architecture for our experiments, details of which are explained in Section~\ref{sec:exp}. For both models, the input is formed by concatenating \texttt{<instruction>} and \texttt{<source>} with a newline, while the output is \texttt{<target>}.

 %We leverage human preference and synthetic preference datasets to train our reward models. The synthetic preference dataset is generated as follows:

% \begin{itemize}

% \item \textbf{\lm{Sampling responses}}. We sample \texttt{<instruction>-<source>} pairs from the synthetic instruction dataset (Section~\ref{section:SyntheticTrainingData}) and obtain outputs \texttt{<target>} from both $M_\text{SFT}$ and $M_\text{base}$.

% \item \textbf{\lm{Ranking}}. To rank the outputs of the models, we employ the methodology outlined in Section~\ref{subsubsection:Post-processing}. We utilize NLI scores to assess content preservation and the degree of hallucination in the generated text. Additionally, we calculate the edit distance between \texttt{<source>} and \texttt{<target>}, considering small edit distances as penalties. Furthermore, we consider the length ratio between \texttt{<target>} and \texttt{<source>} to determine whether there has been an appropriate text expansion or compression. Specifically, for tasks that require adding new information, the length ratio should be greater than $1$, whereas for tasks that emphasize conciseness, the length ratio should be less than $1$. The target with a higher rank is labeled as the ``good target'' ($r_\text{good}$), while the other is referred to as the ``bad target'' ($r_\text{bad}$).

% \end{itemize}
\textbf{Reward Modeling (RM)}
Firstly, we sample prompt data (instruction and source) from our training dataset, and sample outputs from the pretrained language model $M_\text{base}$ and finetuned model $M_\text{SFT}$.

Secondly, in constrast to InstructGPT, where human labelers are used to rank the outputs, we develop a new approach to rank model outputs without any human effort for collecting preference data for reward model training. We define a new scoring function to measure the quality of the rewrite transformation through several heuristics (see Section~\ref{subsubsection:Post-processing}). For an input output pair $(x, t)$, the quality score is defined as follows: 

\begin{equation*}
\small
\centering
Q(x, t) =
\begin{cases}
  0, & \begin{aligned}
      &\text{if EditRatio}(x, t) < a \text{ or} \\
      &\text{NLI}(x, t) < b \text{ or} \\
      &\text{NLI}(t, x) < c \text{ or} \\
      &(I_{\text{shorten}} \& \text{ LenRatio}(x, t) > d_1)
      \end{aligned} \\
      \\
  0, & \begin{aligned}
      &\text{if EditRatio}(x, t) < a \text{ or} \\
      &\text{NLI}(x, t) < b \text{ or} \\
      &\text{NLI}(t, x) < c \text{ or} \\
      &(I_{\text{elaborate}} \& \text{ LenRatio}(x, t) < d_2)
      \end{aligned} \\
      \\
  1, & \text{otherwise}
\end{cases}
\end{equation*}
% \begin{equation}
% \small
%     Q(x, t) =
%     \begin{cases}
%       0, & \text{if}\ \text{EditRatio}(x, t) < a \ \text{or}\ \text{NLI}(x, t) < b \ \text{or}\ \text{NLI}(t, x) < c \ \text{or}\ (I_{shorten} \& \text{ LenRatio}(x, t) > d_1) \\
%       0, & \text{if}\ \text{EditRatio}(x, t) < a \ \text{or}\ \text{NLI}(x, t) < b \ \text{or}\ \text{NLI}(t, x) < c \ \text{or}\ (I_{elaborate } \&  \text{ LenRatio}(x, t) < d_2)} \\
%       1, & \text{otherwise}
%     \end{cases}
% \end{equation}
, where $a = 1.2$, $b = 0.7$, $c = 0.7$, $d_1=0.6$, and $d_2=2$. $I_{shorten}$ means a shorten task, and $I_{elaborate}$ means a expanding or elaboration task. These are decided simply based on keyword matches. If a \( (x, t) \) pair fails to meet any of the heuristic rules, it is assigned a quality score of \( 0 \); otherwise, a score of \( 1 \) is given. If the model outputs from the same prompt are all good or are all bad, we will discard the example. If some outputs are good and some are bad, we will select the top-ranked ones (based on probability in top-p or top-k sampling) from good outputs and bad outputs respectively. 

Finally, we finetune a pre-trained reward model $R_\text{base}$ using the comparison data collected above. This is different from InstructGPT \citep{ouyang2022training}, which trains the reward model from scratch after obtaining a supervised tuned model. Since $R_\text{base}$ is pretrained on general-purpose preference data and not specialized for open-ended rewriting, additional fine-tuning is crucial.

The reward model, denoted as $r_\theta$, employs a transformer-based architecture with a linear regression output layer. It is trained with $t_{\text{good}}$ and $t_{\text{bad}}$ which represent the good and bad targets respectively. The training loss function for the reward model is the entropy of the normalized score difference between the good and bad targets.

\begin{equation}
\small
% \text{loss}(\theta) = - \mathop{\mathbb{E}}_{(x,t_{\text{good}}, t_{\text{bad}} ) \sim D} \log\Big(\sigma\big(r_\theta(x, t_{\text{good}})-r_\theta(x, t_{\text{bad}})\big)\Big)
\text{loss}(\theta) = -\!\mathop{\mathbb{E}}_{(x,t_{\text{good}}, t_{\text{bad}}) \sim D}\!\log\Big(\sigma\big(r_\theta(x, t_{\text{good}})\!-\!r_\theta(x, t_{\text{bad}})\big)\Big)
\end{equation}
% In contrast to InstructGPT \citep{ouyang2022training}, which trains the reward model from scratch after obtaining a supervised tuned model, our approach uses a pre-trained reward model  on general-purpose preference data. Specifically, we use a PaLM 2 (S) reward model pre-trained on human-rated preference data as our base reward model, $R_\text{base}$.
% However, as $R_\text{base}$ is not optimized for open-ended rewriting scenarios, we further fine-tune the model with the following steps:

%We We now create a preference dataset $D_\text{Pref}$ and fine-tune the base reward model $R_\text{base}$ to obtain $R_\text{SFT}$. 

%The input consists of concatenated \texttt{<instruction>}, \texttt{<source>}, and \texttt{<target>} tokens, while the output is a scalar score. We use a good target
%\jd{How do we get good targets and bad targets?} \ls{done, in heuristic ranking}
%$r_\text{good}$ %and a bad target $r_\text{bad}$ for comparison during training. The reward model scores both inputs, and computes the predicted score difference $\hat{r} = \sigma(r_\text{good} - r_\text{bad})$. The training loss is the entropy of the normalized score difference $\hat{r}$.
% \yz{I think the description is not right? Should be cross entropy of ($r_\text{good} - r_\text{bad}$, $r_\text{labels}$), where $r_\text{labels}$ indicates the which input is better?} \ls{since label is always 1, thus liangchen follows instructgpt description}

\textbf{Reinforcement Learning}. Finally, we further optimize the supervised fine-tuned model $M_\text{SFT}$ by employing reinforcement learning~\citep{ouyang2022training}, guided by the scores provided by the fine-tuned reward model $R_\text{SFT}$. This process results in the final model, $M_\text{rewrite}$. % \yz{I think we should at least cite some paper here, indicating the alrogithm we used? We may not able to disclose the ALP algorithm, but maybe cite some others?}\ls{add citation to instructgpt}

\section{Evaluation Framework}
\subsection{OpenRewriteEval --- A New Benchmark for Text Rewriting}

% \begin{figure}[h]
%     \centering
%     \includegraphics[width=0.48\textwidth]{figures/llm_wiki_instruction_plot.png}
%     \includegraphics[width=0.48\textwidth]{figures/openrewriteeval_instruction_plot.png}
%     % \subfloat[Wiki Instructions]{\includegraphics[width=0.33\textwidth]{figures/wiki_instruction_plot.png}}
%     % \subfloat[Synthetic Instructions]{\includegraphics[width=0.33\textwidth]{figures/llm_instruction_plot.png}}
%     \caption{Plotting the instruction (verb+noun) distribution from the training data and testing data (OpenRewriteEval).}
% \end{figure}

% \begin{figure}[h]
% \includegraphics[width=0.98\textwidth]{figures/newplot.png}
% \centering
% \caption{The plot of instructions in \textsc{OpenRewriteEval}. %\lm{if no space, move to appendix}
% }
% \label{fig:instruction-plot}
% \end{figure}

\begin{table*}[h]
\centering
\fontsize{9pt}{9pt}\selectfont
\setlength{\tabcolsep}{1.2pt}
\begin{tabular}{lcccccccccc}
\toprule
           &        &          &          &          &           &           &            &  & \multicolumn{2}{c}{NLI} \\ \cmidrule{10-11} 
           & Size   & Inst Len & Src Len  & Tar Len   & Len Ratio & Edit Dist & Edit Ratio & Rouge1 & src-tar    & tar-src    \\ \midrule
All        & 1629 &6.40&132.71&143.53&1.12&90.79&0.71&67.19 & 0.94 & 0.95 \\

$D_\text{Formality}$  & 200  &5.10  &114.73 &119.23 &1.12   &62.51  &0.56   &68.93 & 0.87 & 0.98 \\
$D_\text{Paraphrase}$ & 102  & 3    & 211.02 & 195.97 & 1    & 121.2  & 0.54 & 68.57 & 1    & 1    \\
$D_\text{Shorten}$    & 102  & 4.49 & 211.02 & 165.68 & 0.8  & 72.2   & 0.37 & 79.26 & 1    & 1    \\
$D_\text{Elaborate}$  & 102  & 8.64 & 211.02 & 378.47 & 2.07 & 234.33 & 1.34 & 56.52 & 0.92 & 1    \\
$D_\text{MixedWiki}$       & 606  & 7.54 & 103.3  & 97.57  & 0.98 & 65.36  & 0.64 & 71.86 & 0.94 & 0.92 \\
$D_\text{MixedOthers}$       & 517  & 6.17 & 127.8  & 145.74 & 1.18 & 100.89 & 0.82 & 60.51 & 0.95 & 0.95 \\
% All        & $1629$ & $6.39$   & $129.02$ & $141.35$ & $1.14$    & $90.34$   & $0.72$     & $0.94$     & $0.95$     \\ 
% $D_\text{Formality}$  & $200$  & $5.04$   & $84.62$  & $101.47$ & $1.28$    & $58.83$   & $0.70$     & $0.87$     & $0.98$     \\
% $D_\text{Paraphrase}$ & $102$  & $3.00$   & $211.02$ & $195.97$ & $1.00$    & $121.20$  & $0.54$     & $1.00$     & $1.00$     \\
% $D_\text{Shorten}$    & $102$  & $4.49$   & $211.02$ & $165.68$ & $0.80$    & $72.20$   & $0.37$     & $1.00$     & $1.00$     \\
% $D_\text{Elaborate}$  & $102$  & $8.64$   & $211.02$ & $378.47$ & $2.07$    & $234.33$  & $1.34$     & $0.92$     & $1.00$     \\
% $D_\text{MixedWiki}$       & $606$  & $7.54$   & $103.30$ & $97.57$  & $0.98$    & $65.36$   & $0.64$     & $0.94$     & $0.92$     \\
% $D_\text{MixedOthers}$      & $517$  & $6.17$   & $127.80$ & $145.74$ & $1.18$    & $100.89$  & $0.82$     & $0.95$     & $0.95$     \\
\bottomrule
\end{tabular}
\caption{
Statistics of \textsc{OpenRewriteEval} 
%Size, Inst Len, Src Len, Tar Len, Len Ratio, Edit Dist, Edit Ratio are introduced in Table~\ref{tab:training-data-statistic}.
the number of examples (Size); the average number of words in instructions (Inst Len), source texts (Src Len), and target texts (Tar Len); the average length fraction (Len Ratio) between the target and source texts; the average edit distance (Edit Dist) between source and target; the ratio of edit distance to source text length (Edit Ratio); and the Rouge1 score comparing source and target texts for the full set and the subtasks. All are measured at the word-level. 
NLI (src-tar, tar-src) are the NLI scores between the source text and the gold reference.
	}
\label{tab:data}
\end{table*}

%\subsection{Open-ended Rewrite Benchmark}
To facilitate the evaluation of open-ended rewriting, we have curated a new dataset called \textsc{OpenRewriteEval}, which focuses on open instructions, long-form text, and large edits. Each example in the dataset consists of a three-tuple $(\texttt{<instruction>}, \texttt{<source>}, \texttt{<target>})$.

\textsc{OpenRewriteEval} consists of six datasets  $D_\text{Formality}$, $D_\text{Paraphrase}$, $D_\text{Shorten}$, $D_\text{Elaborate}$, $D_\text{MixedWiki}$ and $D_\text{MixedOthers}$. See Table \ref{tab:instruction_sample} and Figure~\ref{fig:instruction-plot} for more details about dataset size, data source, and instruction examples. For $D_\text{Formality}$, $D_\text{Paraphrase}$, and $D_\text{Shorten}$, we  use a fixed set of instruction. For the rest of the datasets, we asked human annotators to attach appropriate instructions to each source text and then rewrite them accordingly.
 %\lm{move?} %Also, Table \ref{tab:data} has more detailed stats about each dataset.
% To ensure the quality of human rewrite, we have two kinds of annotation tasks (1) rewrite based on the instruction and (2) rate the rewrite quality. The detailed annotation guidelines are detailed in Appendix~\ref{sec:appendix_guideline}. We remove the low-quality rewrites.
Table \ref{tab:data} provides information on the size of each task and the average word-level lengths of instructions, source text, and target text.
\textsc{OpenRewriteEval} captures how people naturally rewrite, which usually include changes across multiple sentences. This sets us apart from existing benchmarks such as EditEval~\cite{dwivedi-edit-2022}, which are limited to rewrites within single sentences. See Edit ratio (dividing the edit distance by the length of the source text): \textsc{OpenRewriteEval} (0.37-1.34; see Table~\ref{tab:data}) vs EditEval (0.17-0.59; see Table~\ref{tab:editeval_data}).
Appendix~\ref{sec:appendix_guideline_rewrite} provides detailed guidelines for the rewrite annotations.
%\lm{move?} 
% Appendix~\ref{sec:appendix_editeval} Table \ref{tab:data} provides information on the size of each task and the average word-level lengths of instructions, source texts, and target texts. Our source texts are long-form, and the length ratio (Len Ratio) represents the average and standard deviation of the target's length compared to the source text's length. The formality, paraphrase, and open-ended tasks have average length ratios around 1, indicating similar lengths between the source and target texts. The elaborate task aims to expand the source text, resulting in a higher length ratio. Conversely, the shorten task aims to make the source text more concise, resulting in a length ratio smaller than 1. The edit distance metric shows the word-level edits between the source and target texts. We report the fraction of edit distance over the source text's length (edits over source). Most tasks exhibit edit ratios higher than 50\%, indicating that our benchmark dataset involves substantial edits.

\subsection{Automatic Evaluation Metrics}
\label{automatic_metrics}
% We employ various metrics to evaluate the model's performance including Natural Language Inference (NLI) scores, edit distance ratio (Edit Ratio), length ratio (Len Ratio) for the source-prediction pair, SARI, BLEU, GLEU, ROUGE, and Update-ROUGE. For detailed descriptions of these metrics, please refer to Appendix~\ref{sec:appendix_metrics}. \lm{need to add back if we have enough space. Metrics are important}
We employ various metrics to evaluate the model's performance including
\begin{itemize}
  \item \textbf{NLI} \citep{bowman2015large} and \textbf{Reversed NLI} (\ie reverse the premise and the hypotheses) score over the source-prediction pair. NLI and Reversed NLI scores illustrate the model prediction's content presentation and factuality quality. We use the off-the-shelf NLI predictor introduced by \citep{honovich2022true}.
  \item \textbf{Edit Distance Raito (Edit Ratio)}. Edit distance \citep{ristad1998learning} measures the word-level textural difference between two pieces of text. We report the relative edit distance between the prediction and source text, \ie dividing the edit distance by the length of the source text. The edit ratio represents the proportion of the source text that has been modified. It is undesirable if the edit distance is small because this indicates the prediction is primarily identical to the source text. Ideally, we expect to see this value to be neither excessively high (indicating the entire content has been changed) nor excessively low (indicating that only minor rewriting occurred thereby diminishing the perceived effectiveness of the system).
%   \item \textbf{Length ratio (Len Ratio)} over source text computes length fraction between the prediction and the source text. 
  \item \textbf{SARI}~\citep{xu2016optimizingsari} is an n-gram based metric measures how a close a prediction is relative to the source text and the reference text by rewarding words added, kept, or deleted. SARI computes the arithmetic mean of n-gram F1-scores for each of the three operations.

\item \textbf{GLEU} \citep{napoles2015ground} measures the precision of the n-grams in the model's prediction that match the reference.
%is an n-gram based metric that emphasizes the high precision of n-grams between the prediction and the references. 
It is a variant of BLEU~\citep{papineni2002bleu}. GLEU is customized to penalize only the changed n-grams in the targets, as unmodified words do not necessarily need to be penalized in the rewriting task.
\item \textbf{Update-ROUGE (Updated-R)} \citep{iv2022fruit} measures the recall of n-grams between the model's prediction and the references. It is a modified version of ROUGE \citep{lin2003automaticrouge}. Updated-R specifically computes ROUGE-L on the updated sentences rather than the full text.

\end{itemize}

When evaluating quality, it is desirable to have a higher value of NLI. Additionally, a higher Edit Ratio within a reasonable range is preferred. However, it's important to note that considering these metrics independently is insufficient. In some cases, predictions with a low edit ratio may still have high NLI scores. Conversely, a large edit ratio can contain hallucinations if the NLI scores are low.
% The appropriate value for the length ratio depends on the specific rewrite instruction. For instance, tasks that involve shortening text require a lower length ratio, while tasks that require elaboration demand a higher length ratio.
Additionally, higher values of SARI, GLEU, and Update-ROUGE indicate that the predictions are more similar to the gold reference text.

\subsection{Human Evaluation}
We conduct human evaluation on randomly selected 80 examples from the \textsc{OpenRewriteEval} dataset with five language experts. The rating use a 3-point Likert scale (0-Bad, 1-Medium, or 2-Good) for the following features: 1) \textbf{Instruction Success:} whether the rewrite accurately follows the instruction provided. 2) \textbf{Content Preservation:} whether the rewritten text preserves the essential content and meaning of the source text, regardless of its writing style or quality. 3) \textbf{Factuality:} Checks the accuracy and truthfulness of the answer's content. 4) \textbf{Coherency:} whether the rewritten text is easy to understand, non-ambiguous, and logically coherent when read by itself (without checking against the source text). 5) \textbf{Fluency:} Examines the clarity, grammar, and style of the written answer. The detailed rating guideline is in Appendix~\ref{sec:appendix_guideline_rate}.

\section{Experiments and Results}
\label{sec:exp}
This section provides an overview of our experimental settings, baselines, and result analysis. Detailed information about the hyperparameters can be found in Appendix~\ref{sec:appendix_hyperparameter}.

\begin{table*}[h]
\fontsize{9pt}{9pt}\selectfont
\centering
\setlength{\tabcolsep}{4pt}
\begin{tabular}{lrrrrrrr}
\toprule
                                & \multicolumn{1}{l}{} & Edit Ratio & NLI (s-p) & NLI (p-s) & SARI    & GLEU   & Update-R \\ \hline
\multicolumn{8}{l}{\cellcolor[HTML]{DDDDDD}Pretrained LLMs}      \\
% PaLM \citep{chowdhery2022palm}                           & 8B                   & $0.27$     & $0.30$    & $0.12$    & $26.13$ & $0.62$ & $8.62$   \\
PaLM \citep{chowdhery2022palm}                             & 62B                  & $0.31$     & $0.25$    & $0.11$    & $28.24$ & $0.74$ & $11.99$  \\
PaLM 2 \citep{palm2}                        & S                    & \textbf{1.22}     & $0.63$    & $0.37$  & $28.62$ & $0.48$ & $8.14$   \\
LLaMA \citep{touvron2023llama}                          & 65B                  & $0.71$     & $0.83$    & $0.83$   & $27.98$ & $2.10$ & $21.35$  \\ \hline
\multicolumn{8}{l}{\cellcolor[HTML]{DDDDDD}Instruction-Tuned LLMs}      \\
% Alpaca \citep{alpaca}                         & 7B                   & $0.11$     & $0.90$    & $0.85$   & $35.37$ & $5.97$ & $34.14$  \\
Alpaca \citep{alpaca}                           & 13B                  & $0.11$     & $0.90$    & $0.85$  & $36.12$ & $6.81$ & $34.88$  \\
Alpaca-PaLM 2 & S & $0.12$ & $0.9$ & $0.84$ & $38.51$ & $8.31$ & $36.56$ \\

% Vicuna \citep{vicuna2023}                         & 7B                   & $0.22$     & $0.87$    & $0.75$   & $38.48$ & $6.44$ & $32.58$  \\
 Vicuna \citep{vicuna2023}                          & 13B                  & $0.23$     & $0.89$    & $0.77$   & $39.05$ & $6.84$ & $33.31$  \\ 
Flan-PaLM \citep{chung2022scaling}                      & 62B                  & $0.12$     & $0.58$    & $0.42$   & $24.52$ & $1.87$ & $6.23$   \\

InsGPT (text davinci 001) & - & $0.09$ & $0.66$ & $0.61$ & $27.17$ & $3.72$ & $18.69$ \\
ChatGPT (GPT 3.5 Turbo) & - & $0.13$ & $0.95$ & $0.87$ & $40.04$ & $8.47$ & $37.78$ \\
\hline

\multicolumn{8}{l}{\cellcolor[HTML]{DDDDDD}RewriteLMs}      \\
Rewrite-PaLM                  & 62B                  & $0.14$     & $0.88$    & $0.76$   & $37.02$ & $7.40$ & $36.68$  \\
% Rewrite-Flan-PaLM            & 62B                  & $0.15$     & $0.88$    & $0.72$   & $37.74$ & $7.58$ & $37.06$  \\
Rewrite-PaLM 2                  & S                    & $0.25$     & $0.93$    & $0.79$   & $40.92$ & \textbf{9.64} & $39.36$  \\ 
Rewrite-RL-PaLM 2 & S                    & $0.27$     & $0.94$    & $0.81$   & \textbf{40.97} & $9.43$ & $39.36$  \\
Rewrite-RL$_\text{r/w}$-PaLM 2 & S                    & $0.29$     & \textbf{0.96}    & \textbf{0.87}  & $40.66$ & \textbf{9.64} & \textbf{40.10} \\
\bottomrule
\end{tabular}
\caption{Model Performance on \textsc{OpenRewriteEval}. Edit distance ratio (Edit Ratio) between the model prediction and the source text; NLI score with source as premise and model prediction as hypothesis (NLI s-p) and vice versa (NLI p-s); SARI, GLEU and Updated-ROUGE (Updated-R) between the gold reference and the model prediction are reported here.}
\label{tab:perf_auto_all}
\end{table*}

\begin{table*}[h]
\centering
\fontsize{9pt}{9pt}\selectfont
\setlength{\tabcolsep}{4pt}
\begin{tabular}{lcccccccccc}
\toprule
                              &  & \multicolumn{2}{c}{JFL}    & TRK   & AST   & WNC   & \multicolumn{2}{c}{FRU} & \multicolumn{2}{c}{WFI} \\ \cmidrule{3-4} \cmidrule{8-9} \cmidrule{10-11} 
                              &      & SARI       & GLEU         & SARI  & SARI  & SARI  & SARI      & Update-R    & SARI     & Updated-R    \\ \midrule
Copy                           & -    & 26.7          & 40.5          & 26.3           & 20.7           & 31.9          & 29.8          & 0              & 33.6          & -              \\ \midrule
Tk~\citep{wang2022benchmarking-tk}                           & 3B   & 31.8          & 39            & 32.8           & 29.9           & 31.3          & 12.6          & 3.6            & 1.3           & 4.5            \\
T0~\citep{sanh2022multitask}                            & 3B   & 42            & 38.8          & 34.4           & 32.3           & 22.3          & 14.2          & 9.6            & 5.1           & 16.3           \\
T0++~\citep{sanh2022multitask}                          & 11B  & 34.7          & 43.2          & 32.9           & 28.2           & 29.3          & 12.6          & 3.7            & 4.4           & 8.1            \\
PEER-3~\citep{schick2022peer}                         & 3B   & 55.5          & 54.3          & 32.5           & 30.5           & 53.3          & 39.1          & 30.9           & 34.4          & 18.7           \\
PEER-11~\citep{schick2022peer}                        & 11B  & 55.8          & 54.3          & 32.1           & 29.5           & \textbf{54.5} & \textbf{39.6} & 31.4           & \textbf{34.9} & 20.4           \\
OPT~\citep{zhang2022opt}                            & 175B & 47.3          & 47.5          & 32.6           & 31.8           & 31.2          & 35.9          & 27.3           & 26.7          & 11.2           \\
GPT-3~\citep{brown2020language-gpt3}                          & 175B & 50.3          & 51.8          & 33             & 30.5           & 31.7          & 36            & 21.5           & 27.2          & 10.6           \\
InsGPT~\citep{ouyang2022training}                         & 175B & \textbf{61.8} & \textbf{59.3} & 38.8           & 38             & 35.4          & 36.3          & 24.7           & 23.6          & 16.1           \\ \midrule
PaLM 2~\citep{palm2}                    & S    & 36.07         & 2.18          & 34.32          & 35.92          & 25.2          & 24.28         & 26.39          & 11.41         & 20.42          \\
Rewrite-PaLM 2 (Ours)                & S    & 56.95         & 40.38         & 40.81          & \textbf{42.11} & 37.11         & 37.51         & 53.54          & 26.55         & 47.06          \\
Rewrite-RL$_\text{r/w}$-PaLM 2 (Ours) & S    & 55            & 22.89         & \textbf{40.87} & 41.71          & 37.81         & 38.56         & \textbf{53.93} & 29.25         & \textbf{49.53} \\
\bottomrule
\end{tabular}
\caption{Model Performance on EditEval~\citep{dwivedi-edit-2022}.}
\label{tab:perf_editeval}
\end{table*}
\subsection{Baselines}
\label{subsec:baselines_and_metrics}

\begin{table*}[t]
\centering
\small
\setlength{\tabcolsep}{2pt}
\begin{tabular}{l|p{1.3cm}p{1.6cm}lll|l}
\toprule
& Instruction Success & Content Preservation & Factuality & Coherence & Fluency & AVG \\
\midrule
Agreement           & 0.784 & 0.781 & 0.769 & 0.933 & 0.804 & 0.814 \\ \midrule
Human Expert & 1.833 & 1.949 & 1.985 & 1.99  & 1.99  & 1.949 \\\midrule
Alpaca 13B       & 1.441 & 1.754 & 1.934 & 1.962 & 1.977 & 1.814 \\
Alpaca-PaLM 2     & 1.489 & 1.719 & \textbf{1.99}  & \textbf{2}     & \textbf{2}     & 1.839 \\
ChatGPT          & 1.478 & 1.775 & 1.959 & 1.962 & 1.975 & 1.83  \\
Rewrite-PaLM 2    & 1.641 & 1.777 & 1.927 & \textbf{2}     & \textbf{2}     & 1.869 \\
Rewrite-RL$_\text{r/w}$-PaLM 2 & \textbf{1.648} & \textbf{1.835} & 1.959 & 1.985 & \textbf{2}     & \textbf{1.886} \\
\bottomrule
\end{tabular}
\caption{Human Evaluation Results.}
\label{tab:humaneval_results}
\end{table*}

We use the following baseline models for quality comparison in the later sections:

\begin{itemize}
\item \textbf{PaLM}~\citep{chowdhery2022palm} is a large, densely activated transformer-based language model that can generate text in an open-ended fashion.
\item \textbf{PaLM 2}~\citep{palm2}, is an advanced language model which surpasses its predecessor PaLM in terms of multilingual and reasoning abilities while being more computationally efficient. It is a Transformer-based model that underwent training using a blend of objectives. In this paper, we employ PaLM 2-S. This ``S" size is comparable to LLaMA/Alpaca/Vicuna-13B, which is why we opted to train using it rather than the largest PaLM 2. Note that the specific number of parameters for the PaLM 2 series has not been made public. Instead, the PaLM 2 Tech Report uses T-shirt sizes to represent model sizes, ranging from XXS to L. We follow its notations.
\item \textbf{LLaMA}~\citep{touvron2023llama} is an efficient, open-source foundation language model.
\item \textbf{Flan-PaLM}~\citep{chung2022scaling} is fine-tuned on a large variety of tasks and chain-of-thought data using PaLM as the base model.
\item \textbf{Alpaca}~\citep{alpaca} is a language model that is fine-tuned from LLaMA using 52,000 instruction-following demonstrations.
\item \textbf{Alpaca-PaLM:} We fine-tune the PaLM model on Alpaca instruction-following datasets.
\item \textbf{Vicuna}~\citep{vicuna2023} is an open-source chatbot trained by fine-tuning LLaMA on user-shared conversations collected from ShareGPT\footnote{https://sharegpt.com/}.
\item  \textbf{ChatGPT} (gpt-3.5-turbo)\footnote{\url{https://openai.com/chatgpt}} and  \textbf{InsGPT} (text-davinci-001)~\cite{ouyang2022training} are members of the GPT family, developed by OpenAI.
\end{itemize}
We follow the same zero-shot prompt setting for all the baseline models as \citet{schick2022peer}. The pre-trained models without any instruction tuning generally exhibit slightly lower performance in following instructions compared to the instruction tuned models under zero-shot scenario. 

\subsection{Results on \textsc{OpenRewriteEval} Benchmark}

\label{subsec:resultanalysis}
The automatic evaluation results for the \textsc{OpenRewriteEval} dataset are presented in Table~\ref{tab:perf_auto_all}. Rewrite-PaLM and Rewrite-PaLM 2 are supervised fine-tuned versions (as discussed in Section \ref{training_procedure}) based on PaLM, and PaLM 2, respectively. Rewrite-RL-PaLM 2 and Rewrite-RL$_\text{r/w}$-PaLM 2 are reinforcement learning models tuned over Rewrite-PaLM 2. The reward model from the former does not use our synthetic preference dataset (as discussed in Section \ref{training_procedure}), whereas the reward model from the latter incorporates it.

As shown in Table~\ref{tab:perf_auto_all}, our RL tuned model Rewrite-RL$_\text{r/w}$-PaLM 2 has the highest scores in almost all the metrics (i.e., NLI scores, SARI, GLEU, and Update-R). This indicates that our model is good at generating outputs faithful to the original input, while other models might generate more ``hallucinations''. For edit ratio, Rewrite-RL$_\text{r/w}$-PaLM 2 has a better score than all the models except PaLM 2. Pre-trained models such as PaLM 2 without any instruction tuning are prone to generating ``hallucinations'', resulting in a significantly high edit ratio score (\ie $1.22$). Therefore, our model is good at keeping all the essential content and meaning of the source text, while also being able to rewrite with varied language and structures. Given that Rewrite-RL$_\text{r/w}$-PaLM~2 consistently outperforms Rewrite-RL-PaLM~2 across nearly all metrics, this strongly suggests the effectiveness and value of employing synthetic preference data.

See Appendix \ref{sec:appendix_results} for more metrics on OpenRewriteEval dataset (see Table~\ref{tab:openrewriteeval_full}) and a breakdown by each subgroup (see Tables \ref{tab:openrewriteeval_formality}, \ref{tab:openrewriteeval_paraphrase}, \ref{tab:openrewriteeval_shorten}, \ref{tab:openrewriteeval_elaborate}, \ref{tab:openrewriteeval_wiki}, \ref{tab:openrewriteeval_others}).

\subsection{Results on EditEval}

We also evaluated the performance of our models using the publicly available sentence-level rewrite benchmark EditEval\footnote{\url{https://github.com/facebookresearch/EditEval}} \citep{dwivedi-edit-2022}. This benchmark comprises various datasets that cover different language tasks. Specifically, JFL ~\citep{napoles2017gec} focuses on language fluency; TRK~\citep{xu2016optimizingsari} and AST \citep{alva-manchego-etal-2020-asset} target at sentence simplification;  WNC \citep{pryzant2020automatically-wnc} addresses text neutralization; FRU~\citep{iv2022fruit}) and WFI ~\citep{petroni2022improving-wafer} involve updating information that requires external references. More  data statistics for each dataset can be found in Table~\ref{tab:editeval_data}.

We only report the results on EditEval datasets that containing more than 100 test examples (see Table~\ref{tab:perf_editeval}). The results of LLM baselines and the Copy baseline (which treats the source text as the prediction) are taken directly from the EditEval paper~\citep{dwivedi-edit-2022}. We can observe that the zero-shot performance of Rewrite-PaLM 2 and Rewrite-RL$_\text{r/w}$-PaLM 2 is mostly on par with or better than the best baselines (\ie PEER-11 and InsGPT). While our model is specifically designed for long-form text rewriting, it does not sacrifice its capability to handle sentence-level rewriting tasks.

\subsection{Results on Human Evaluation}
\label{sec:exp_human_eval}

% \begin{figure}[h]
% \includegraphics[width=0.5\textwidth]{figures/rewriteplam2_wintielose_plot.png}
% \centering
% \caption{Human preference evaluation, comparing Rewrite-PaLM 2 to 5 different baselines over 80 examples. }
% \label{fig:wintielose}
% \end{figure}

The human evaluation results, detailed in Table~\ref{tab:humaneval_results}, reveal notable insights. The inter-annotator agreements, quantified using the Fleiss kappa coefficient~\cite{fleiss1971measuring}, underscore the reliability of the evaluations. Notably, Rewrite-PaLM 2 and Rewrite-RW$_\text{r/w}$-PaLM 2 demonstrate superior performance over Alpaca, Alpaca-PaLM 2, and ChatGPT in instruction success and content preservation. This alignment with the automatic evaluation metrics underscores the efficacy of these models in adhering to given instructions while maintaining the integrity of the original content.
In terms of coherence and fluency, all models, including the human rewrites, scored above 1.96, indicative of their ability to generate clear, unambiguous, and logically coherent outputs. Such high scores suggest that these models' outputs are not only understandable but also align closely with human-level language proficiency. Human expertise still prevails in aspects of instruction success and content preservation, suggesting room for further improvement in model performance to reach human-level proficiency in rewriting tasks.

\section{Conclusion}
% This paper introduces a novel benchmark for text rewriting that overcomes
% \jd{Can a benchmark "overcomes" a challenge? To me, a benchmark probably quantifies a challenge and making it measurable. A benchmark may not be able to enable rewriting either. Can we rephrase to "enable research in rewriting" or something like that? }
% the challenges of unintended modifications and enables open-ended rewriting of long-form text. To support our benchmark, we propose RewriteLM, an instruction-tuned large language model for text rewriting. We also present new data generation strategies to reduce human intervention. Our experiments demonstrate the effectiveness of our proposed benchmark, baseline model, and data collection strategies. \lm{update based on results}

We introduce a novel benchmark for text rewriting with a focus on cross-sentence rewrites, covering a wide variety of rewriting types expressed through natural language instructions. We present new data generation and training strategies to better teach LLMs to perform rewriting tasks. Our model, RewriteLM, achieves the state-of-the-art results on \textsc{OpenRewriteEval} benchmark.
\section{Acknowledgments}
The authors would like to thank Tony Mak, Chang Li, Abhanshu Sharma, Matt Sharifi, Hassan Mansoor, Daniel Kim, Reut Aharony and Nevan Wichers for their insightful discussions and support.

\bibliography{custom}

\appendix
\section{Appendix}
\label{sec:appendix}
\subsection{\textsc{OpenRewriteEval} Data}
\label{openrewrite_eval}

\begin{figure*}[h]
    \centering
    \begin{subfigure}[b]{0.48\textwidth}
        \centering
        \includegraphics[width=\textwidth]{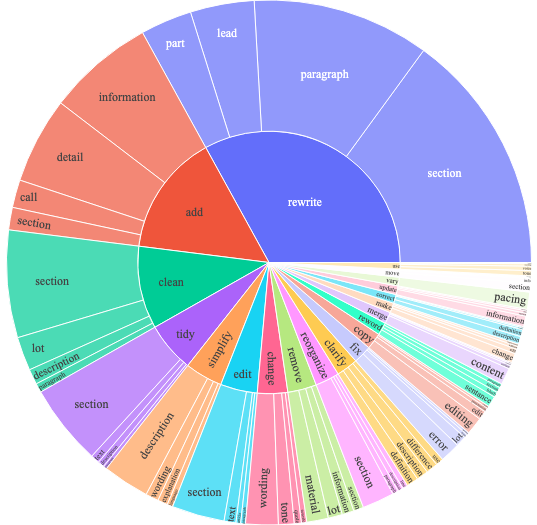}
        \caption{Training Data}
        \label{fig:instruction-plot-training}
    \end{subfigure}
    \hfill % To add some space between the subfigures
    \begin{subfigure}[b]{0.48\textwidth}
        \centering
        \includegraphics[width=\textwidth]{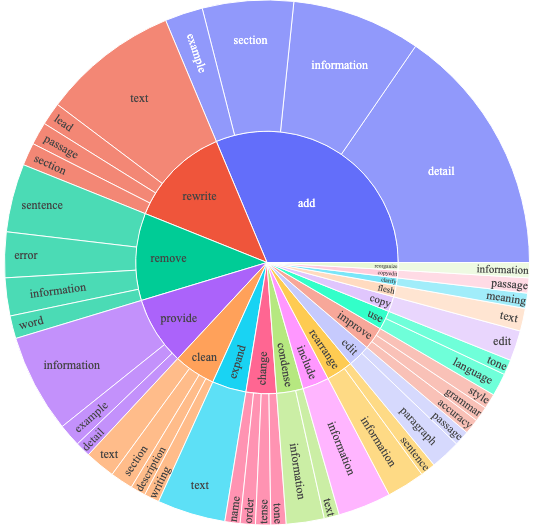}
        \caption{OpenRewriteEval}
        \label{fig:instruction-plot-openrewriteeval}
    \end{subfigure}
    \caption{Plotting the instruction (verb+noun) distribution from the training data and testing data (OpenRewriteEval).}
    \label{fig:instruction-plot}
\end{figure*}

\begin{table*}
\centering
\fontsize{9pt}{9pt}\selectfont
\setlength{\tabcolsep}{1.2pt}
\begin{tabular}{lcll}
\toprule
Dataset & Size & Data Source & \hspace{0.5cm}Instruction Examples\\ 
\midrule

$D_\text{Formality}$           & 200  & See Appendix \ref{openrewrite_eval} &    \hspace{0.5cm}\begin{tabular}[c]{@{}l@{}}Too conversational, rephrase it to be more formal?\\ Make the text more formal.\\ Rephrase it to be more formal?\end{tabular}\\\midrule
$D_\text{Paraphrase}$       & $102$ & See Appendix \ref{openrewrite_eval} &  \hspace{0.5cm}\begin{tabular}[c]{@{}l@{}}Paraphrase this.\\ Reword this text.\\ Use different wording.\end{tabular}                                                                                                                                                                                                                                                               \\\midrule
$D_\text{Shorten}$          & $102$ & See Appendix \ref{openrewrite_eval} & \hspace{0.5cm}\begin{tabular}[c]{@{}l@{}}Make wording more concise.\\ Improve accuracy, clarity, and conciseness of language.\\ Rephrase for clarity and conciseness.\end{tabular}                                                                                                                                                                                                \\\midrule
$D_\text{Elaborate}$        & $102$ & See Appendix \ref{openrewrite_eval} & \hspace{0.5cm}\begin{tabular}[c]{@{}l@{}}Elaborate on advantages of JavaScript.\\ Add more details about fighting styles.\\ Describe more about what the third page does.\end{tabular}                                                                                                                                                                                            \\\midrule
$D_\text{MixedWiki}$  & $606$  & Wiki & \hspace{0.5cm}\begin{tabular}[c]{@{}l@{}}Attempt to make the text sound less like an advertisement.\\ Change to have a consistent past tense throughout the paragraph.\\ Rewrite text in the present tense.\\ Give a detailed and concise description of the Wollyleaf bush.\\ Rewrite for clarity and encyclopedic tone.\end{tabular} \\\midrule
$D_\text{MixedOthers}$ & $517$ & C4, Human & 
\hspace{0.5cm}\begin{tabular}[c]{@{}l@{}}Make it more personal and friendly.\\ Rewrite to haiku.\\ Change the name to Horton Beach throughout the text.\\ Make it more motivational for parents of age 50.\\ Create bullet points from text.\end{tabular}       \\\midrule

All                    & $1629$ & & \\ 

\bottomrule

\end{tabular}
\caption{
The data statistics and instruction samples of \textsc{OpenRewriteEval}.	}
\label{tab:instruction_sample}
\end{table*}

\textbf{Frequent Words}. To illustrate the open-ended instruction statistics, we cluster the open-ended instruction by LDA topic model~\citep{blei2003latent}\footnote{
\url{https://radimrehurek.com/gensim/}} and report top-$10$ frequent words in the $10$ topics in the Table~\ref{tab:topic}. Our open-ended instructions cover a wide range of rewrite requirements. Besides the top words, the rewrite benchmark includes surprise instructions like ``haiku'', ``Shakespeare'', Etc.

\textbf{Data Sources}. The source texts for the $D_\text{Formality}$, $D_\text{Paraphrase}$, $D_\text{Shorten}$, and $D_\text{Elaborate}$ categories are from various datasets, including Multi-News~\citep{fabbri2019multi}, Wikipedia~\citep{guo2020wiki}, PG-19 book~\citep{raecompressivepg19}, BIGPATENT~\citep{sharma2019bigpatent}, BillSum~\citep{kornilova2019billsum}, government reports~\citep{huang2021efficient}, scientific papers~\citep{cohan2018discourse}, Enron email~\citep{zhang2019email}, Reddit~\citep{hamilton2017inductive}, IMDB, and Yelp reviews~\citep{maas2011learning, zhang2015character}. The $D_\text{MixedWiki}$ have the source texts from Wikipedia~\citep{guo2020wiki} and $D_\text{MixedOthers}$ contains C4~\citep{raffel2020exploringc4} and human written ones.

\begin{table*}[h]
    %\fontsize{9pt}{9pt}\selectfont
	\centering
% 	\small
	\begingroup
	%\setlength{\tabcolsep}{1.5pt}
	%\resizebox{\textwidth}{!}{%
	\begin{tabular}{cl}
	\toprule
Topic& Top-frequent Words\\ \midrule
1 & expand, easy, text, make, clear, sure, understand, idea, post, reader \\\midrule
2 & use, make, concise, active, voice, copy, edit, points, write, polite \\\midrule
3 & technical, elaborate, make, job, accessible, add, details, expand, idea, audience \\\midrule
4 & less, add, formal, tone, table, change, contents, detail, make, sound \\\midrule
5 & make, concise, personal, persuasive, positive, friendly, text, person, list, tone \\\midrule
6 & make, engaging, rewrite, polite, accessible, general, audience, sound, objective, text \\\midrule
7 & add, details, conclusion, action, call, product, headline, job, person, make \\\midrule
8 & write, prose, language, create, points, tone, polite, use, objective, formal \\\midrule
9 & change, add, tense, past, examples, present, statistics, tone, formal, table \\\midrule
10 & write, style, add, section, formal, list, engaging, personal, job, product \\
\bottomrule
	\end{tabular}
	%}
	\endgroup
	\caption{Open-ended Instruction Top-10 words in 10 topics.}
	\label{tab:topic}
\end{table*}

% \subsection{Human Annotation Guideline}
% \label{sec:appendix_guideline}
\subsection{Human Rewrite Guideline}
\label{sec:appendix_guideline_rewrite}

%\lm{no open-ended guideline?}

\begin{itemize}

\item	Raters align source text to the instruction, and then rewrite. If the source text is already met the instruction, for example, ``make it more formal'', then treat the source text as target text and rewrite less formal (put at source side).
\item	Ensure (1) the content preservation between source and rewrite; (2) maximum word change; and (3) source and target texts are well aligned with instruction. For example, if the instruction is to ``make it more polite'', then ensure that the target text is much more polite than the source text.
\item Elaborate: the rewrite matches source text's tone and format. Add more relevant information and ideas, but do not make up facts. 
\item Rephrase: the rewrite matches source text's tone, verbosity, format and max changes to existing words.
\item Shorten: the rewrite matches source text's tone and format, trims unnecessary words, simplifies sentences, makes them more concise.
\item Informal-to-Formal: Rewrite the given paragraph so that it is more formal in style. To make the text more formal, try to:
(1) Replace informal words associated with chatty spoken styles (such as slang and contractions) with more formal vocabulary.
(2) Make the text impersonal: avoid referring directly to the author(s) or reader(s), or expressing subjective opinions.
(3) Use strictly standard grammatical forms.

\item Formal-to-Informal: Rewrite the given paragraph so that it is less formal in style. To make your writing less formal, try to:
(1) Replace long or uncommon words with relaxed, everyday terms. You may include contractions (such as changing ``cannot'' to ``can’t'' if it helps the text flow better.
(2) Where appropriate, identify the author and the reader to make the text more relatable. (For example, you might be able to change ``It is believed that...'' to ``I think tha...'')
(3) If a sentence is very long or stiffly phrased, try breaking it up or rearranging it, even if this doesn’t fit the strictest rules of standard grammar.

\end{itemize}

\subsection{Human Rating Guideline}
\label{sec:appendix_guideline_rate}
% We ask linguistic expert to rate the rewrite quality based on the source text on the following 4 dimensions: content preservation, faithfulness, coherence, and linguistic accuracy. Each dimension is scored on a 3-point scale. 

\textbf{Instruction Success}: The ability of the model to adhere to the given instruction is evaluated in this criterion. It is:
\begin{itemize}
 \item Score 2 (Fully/Mostly Followed): if the model output entirely adheres to the provided instructions, demonstrating a clear understanding and implementation of the given task. Or the output mostly adheres to the instructions, with minor deviations or errors.
 \item Score 1 (Partially Followed): if the model output shows some adherence to the instructions but deviates significantly in certain aspects or fails to completely implement them, leading to partial fulfillment of the task.
 \item Score 0 (Not Followed/Mostly Ignored): if the model output largely ignores the provided instructions, making it evident that the task has not been understood or implemented properly. Or despite some slight adherence, the output largely deviates from the intended task as per the instructions.
\end{itemize}

\textbf{Content Preservation}: The essential content and meaning of the reference is preserved in the rewrite, independent of its style or the quality of the writing. It is: 
\begin{itemize}
 \item Score 2 (Fully/Mostly Preserved): if the rewrite is an excellent representation of the content in the reference, with no omissions. Or the rewrite mostly matches the content of the reference, but one or two elements of the meaning have been lost.
 \item Score 1 (Half Preserved): if some of the content is present in the rewrite but approximately the same amount is missing.
 \item Score 0 (Not Preserved/Mostly Lost): if the rewrite is entirely unrelated to the reference. Or despite some slight similarities, the rewrite is hard to recognize as being based on the reference.
 \end{itemize}

\textbf{Factuality}: The rewrite only provides as much information as is present in the reference, without adding anything. It is not misleading and does not make any false statements (unless these were also present in the reference).

\begin{itemize}
  \item Score 2 (Fully/Mostly faithful): Everything in the rewrite is grounded in the reference. Or the rewrite says something that is not mentioned in the reference or contradicts the reference, but it is not an important addition or it is hard to say whether the statement is true or false.
  \item Score 1 (Partly faithful): The rewrite adds significant factual statements to the reference. These may be inaccurate or otherwise not based on the reference, but do not entirely undermine the faithfulness of the rewrite as a whole.
  \item Score 0 (Not/Slightly faithful): The rewrite is mostly wrong, made up, or contradicts what is in the reference text.
 \end{itemize}

\textbf{Coherence}: The rewrite is coherent if, when read by itself (without checking against the reference), it's easy to understand, non-ambiguous, and logically coherent. On the other hand, the rewrite is not coherent if it's difficult to understand what it is trying to say.

 \begin{itemize}
  \item Score 2 (Good): The whole of the rewrite is mostly fluent and easy to read, independent of any reference content. Some specific parts of the rewrite could be more naturally phrased, but overall it is fairly clear and easy to understand.
  \item Score 1 (Neutral): The rewrite is comprehensible, though not on the first read or only with some effort.
  \item Score 0 (Bad): The rewrite is very hard to understand, except by checking against the reference.
 \end{itemize}
 
\textbf{Fluency}: The rewrite is considered fluent if it follows all the rules of its language, including spelling, grammar and punctuation. It reads as though it was written by someone who speaks English as their first language.

\begin{itemize}
\item  Score 2 (Flawless/Good): The rewrite is grammatically correct, contains no spelling errors, and follows all other linguistic rules. An average English speaker would not see anything that looks “wrong”. Or there are just one or two linguistic errors or non-standard formulations, but nothing serious.

\item  Score 1 (Flawed): The rewrite contains a number of errors of different types, but these errors, even when taken together, do not make the text significantly harder to understand.

\item  Score 0 (Poor): The rewrite contains a large number of errors, so that some sections of the text are hard to understand, but other parts are more manageable.

\end{itemize}

% \subsection{Metrics}
% \label{sec:appendix_metrics}

\subsection{Hyper-parameter Setting}
\label{sec:appendix_hyperparameter}
We use $64$ Tensor Processing Units (TPU) V3 chips for fine-tuning. The batch size is $32$, and the maximum training step is $5000$. We use the Adafactor optimizer~\citep{shazeer2018adafactor} with a learning rate of $0.003$. Both the input and output sequence lengths are set to $1024$ tokens. The training dropout rate is $0.1$. During inference, the temperature is set to $0.5$, and the top-K value is $40$.

\subsection{EditEval Data}
\label{sec:appendix_editeval}
Table~\ref{tab:editeval_data} shows the EditEval \citep{dwivedi-edit-2022} Data statistics.
\begin{table*}[h]
\centering
\fontsize{9pt}{9pt}\selectfont
\setlength{\tabcolsep}{1.2pt}
\begin{tabular}{lrrrrrrrr}
\toprule
 &
  \multicolumn{1}{l}{Size} &
  \multicolumn{1}{l}{Inst Len} &
  \multicolumn{1}{l}{Src Len} &
  \multicolumn{1}{l}{Tar Len} &
  \multicolumn{1}{l}{Len Ratio} &
  \multicolumn{1}{l}{Edit Dist} &
  \multicolumn{1}{l}{Edit Ratio} &
  \multicolumn{1}{l}{Rouge1} \\\midrule
JFL  & 747  & 4.55 & 17     & 17.3   & 1.12 & 5.54  & 0.42 & 84.46 \\
TRK  & 359  & 3.1  & 20.24  & 17.43  & 0.88 & 9.5   & 0.47 & 77.85 \\
AST & 359  & 3.1  & 19.72  & 16.74  & 0.86 & 11.21 & 0.58 & 71.74 \\
WNC    & 1000 & 4.33 & 25.96  & 25.69  & 0.99 & 3.29  & 0.17 & 95.37 \\
FRU  & 914  & 3.33 & 116.86 & 131.02 & 1.3  & 47.24 & 0.59 & 78.16 \\
WFI  & 4565 & 3.33 & 200.13 & 221.88 & 1.14 & 32.08 & 0.18 & 92.36 \\
\bottomrule
\end{tabular}
\caption{EditEval Dataset Statistics. Metrics are the number of examples (Size); the average number of words in instructions (Inst Len), source texts (Src Len), and target texts (Tar Len); the average length fraction (Len Ratio) between the target and source texts; the average edit distance (Edit Dist) between source and target; the ratio of edit distance to source text length (Edit Ratio); and the Rouge1 score comparing source and target texts. All measurements are conducted at the word-level.}
\label{tab:editeval_data}
\end{table*}

\subsection{Additional Experimental Results}
\label{sec:appendix_results}
We present comprehensive results for automatic metrics on the full set and each subtask. Table~\ref{tab:openrewriteeval_full} presents the models' performance on the full set of \textsc{OpenRewriteEval}. Tables~\ref{tab:openrewriteeval_formality}, \ref{tab:openrewriteeval_paraphrase}, \ref{tab:openrewriteeval_shorten}, \ref{tab:openrewriteeval_elaborate}, \ref{tab:openrewriteeval_wiki}, and \ref{tab:openrewriteeval_others} show the performance on the formality, paraphrase, shorten, elaborate, mixed Wiki, and mixed others tasks, respectively.

\begin{table*}[h]
\centering
\scriptsize
% \fontsize{10pt}{10pt}\selectfont
% \setlength{\tabcolsep}{2pt}
\begin{tabular}{lrrrrrrrrr}
\toprule
All                             &           &           & \multicolumn{2}{c}{NLI} &         &         &        & \multicolumn{2}{c}{ROUGE-L} \\\cmidrule{4-5} \cmidrule{9-10}
                                & Edit Ratio & Len Ratio & s-p        & p-s        & SARI    & BLEU    & GLEU   & All          & Updated      \\ \hline
\multicolumn{10}{l}{\cellcolor[HTML]{DDDDDD}Pretrained LLMs}      \\
PaLM-8B                         & $0.27$    & $0.97$    & $0.30$     & $0.12$     & $26.13$ & $2.46$  & $0.62$ & $9.78$       & $8.62$       \\
PaLM-62B                        & $0.31$    & $1.36$    & $0.25$     & $0.11$     & $28.24$ & $2.87$  & $0.74$ & $13.35$      & $11.99$      \\
PaLM 2-S                         & $1.22$    & $5.87$    & $0.63$     & $0.37$     & $28.62$ & $2.07$  & $0.48$ & $8.43$       & $8.14$       \\
LLaMA-65B                       & $0.71$    & $4.28$    & $0.83$     & $0.83$     & $27.98$ & $11.66$ & $2.10$ & $25.72$      & $21.35$      \\ \hline
\multicolumn{10}{l}{\cellcolor[HTML]{DDDDDD}Instruction-Tuned}      \\
Alpaca-7B                       & $0.11$    & $0.90$    & $0.90$     & $0.85$     & $35.37$ & $22.80$ & $5.97$ & $43.40$      & $34.14$      \\
Alpaca-13B                      & $0.11$    & $0.92$    & $0.90$     & $0.85$     & $36.12$ & $23.45$ & $6.81$ & $43.95$      & $34.88$      \\
Alpaca-PaLM-S & $0.12$ & $0.85$ & $0.9$  & $0.84$ & $38.51$ & $20.93$ & $8.31$ & $41.39$ & $36.56$ \\
Vicuna-7B                       & $0.22$    & $1.43$    & $0.87$     & $0.75$     & $38.48$ & $15.72$ & $6.44$ & $34.93$      & $32.58$      \\
Vicuna-13B                      & $0.23$    & $1.50$    & $0.89$     & $0.77$     & $39.05$ & $16.39$ & $6.84$ & $35.79$      & $33.31$      \\
Flan-PaLM-62B                       & $0.12$    & $0.68$    & $0.58$     & $0.42$     & $24.52$ & $13.45$ & $1.87$ & $28.87$      & $6.23$       \\ 
%ChatGPT       & $0.13$ & $0.94$ & $0.95$ & $0.87$ & $40.04$ & $18.71$ & $8.47$ & $39.31$ & $37.78$ \\
InsGPT        & $0.09$ & $0.62$ & $0.66$ & $0.61$ & $27.17$ & $21.83$ & $3.72$ & $36.61$ & $18.69$ \\
\hline
\multicolumn{10}{l}{\cellcolor[HTML]{DDDDDD}RewriteLMs}      \\
Rewrite-PaLM-62B                    & $0.14$    & $1.19$    & $0.88$     & $0.76$     & $37.02$ & $25.63$ & $7.40$ & $46.46$      & $36.68$      \\
Rewrite-Flan-PaLM-62B               & $0.15$    & $1.15$    & $0.88$     & $0.72$     & $37.74$ & $24.54$ & $7.58$ & $45.20$      & $37.06$      \\
Rewrite-PaLM 2-S                 & $0.25$    & $1.61$    & $0.93$     & $0.79$     & $40.92$ & $23.56$ & $9.64$ & $44.06$      & $39.36$      \\ 
Rewrite-RL-PaLM 2-S & $0.27$    & $1.72$    & $0.94$     & $0.81$     & $40.97$ & $23.29$ & $9.43$ & $43.60$      & $39.36$      \\
Rewrite-RL$_\text{r/w}$-PaLM 2-S  & $0.29$    & $1.91$    & $0.96$     & $0.87$     & $40.66$ & $24.55$ & $9.64$ & $44.85$      & $40.10$     \\
\bottomrule
\end{tabular}
\caption{Model Performance on \textsc{OpenRewriteEval} full set.}
\label{tab:openrewriteeval_full}
\end{table*}

\begin{table*}[h]
\centering
\scriptsize
% \fontsize{9pt}{9pt}\selectfont
% \setlength{\tabcolsep}{1.5pt}
\begin{tabular}{lrrrrrrrrr}
\toprule
$D_\text{Formality}$                & \multicolumn{1}{l}{} & \multicolumn{1}{l}{} & \multicolumn{2}{c}{NLI} & \multicolumn{1}{l}{} & \multicolumn{1}{l}{} & \multicolumn{1}{l}{} & \multicolumn{2}{c}{ROUGE-L} \\\cmidrule{4-5} \cmidrule{9-10}
                                & Edit Ratio            & Len Ratio            & s-p        & p-s        & SARI                 & BLEU                 & GLEU                 & All          & Updated      \\ \hline
\multicolumn{10}{l}{\cellcolor[HTML]{DDDDDD}Pretrained LLMs}      \\
PaLM-8B                         & $0.30$               & $0.99$               & $0.29$     & $0.12$     & $23.60$              & $2.74$               & $0.40$               & $8.32$       & $6.97$       \\
PaLM-62B                        & $0.41$               & $1.75$               & $0.24$     & $0.14$     & $27.50$              & $3.50$               & $0.81$               & $14.06$      & $12.11$      \\
PaLM 2-S                         & $1.62$               & $7.56$               & $0.65$     & $0.42$     & $27.40$              & $2.92$               & $0.80$               & $8.85$       & $7.78$       \\
LLaMA-65B                       & $0.97$               & $5.43$               & $0.83$     & $0.84$     & $28.88$              & $11.34$              & $2.57$               & $25.30$      & $22.42$      \\ \hline
\multicolumn{10}{l}{\cellcolor[HTML]{DDDDDD}Instruction-Tuned}      \\
Alpaca-7B                       & $0.09$               & $0.92$               & $0.98$     & $0.90$     & $39.69$              & $23.13$              & $8.75$               & $48.22$      & $42.94$      \\
Alpaca-13B                      & $0.11$               & $0.99$               & $0.98$     & $0.92$     & $41.94$              & $23.52$              & $10.43$              & $48.09$      & $44.70$      \\
Alpaca-PaLM 2-S & $0.12$ & $1.04$ & $0.99$ & $0.96$ & $43.94$ & $22.07$ & $12.43$ & $46.68$ & $45.81$ \\

Vicuna-7B                       & $0.16$               & $1.27$               & $0.93$     & $0.87$     & $41.34$              & $17.79$              & $9.42$               & $40.59$      & $39.65$      \\
Vicuna-13B                      & $0.19$               & $1.47$               & $0.95$     & $0.89$     & $42.04$              & $17.41$              & $9.24$               & $39.57$      & $38.61$      \\
Flan-PaLM-62B                       & $0.04$               & $0.84$               & $0.87$     & $0.81$     & $23.32$              & $30.33$              & $6.34$               & $52.94$      & $5.84$       \\  
%ChatGPT & $0.13$ & $1.23$ & $1$ & $0.99$ & $42.36$ & $18.56$ & $10.12$ & $43.42$ & $42.74$ \\
InsGPT & $0.05$ & $0.86$ & $0.86$ & $0.85$ & $29.65$ & $29.76$ & $6.7$ & $52.27$ & $28.97$ \\
\hline
\multicolumn{10}{l}{\cellcolor[HTML]{DDDDDD}RewriteLMs}      \\
Rewrite-PaLM-62B                    & $0.06$               & $1.00$               & $0.99$     & $0.98$     & $44.80$              & $33.48$              & $14.59$              & $59.19$      & $55.07$      \\
Rewrite-Flan-PaLM-62B               & $0.05$               & $1.00$               & $1.00$     & $0.98$     & $45.63$              & $35.91$              & $15.06$              & $61.50$      & $55.81$      \\
Rewrite-PaLM 2-S                 & $0.07$               & $1.02$               & $0.99$     & $0.99$     & $52.39$              & $37.83$              & $23.08$              & $62.64$      & $60.17$      \\ 
Rewrite-RL-PaLM 2-S & $0.07$               & $1.02$               & $1.00$     & $0.99$     & $53.05$              & $38.22$              & $23.61$              & $62.64$      & $60.19$      \\
Rewrite-RL$_\text{r/w}$-PaLM 2-S  & $0.07$               & $1.04$               & $1.00$     & $0.99$     & $52.42$              & $38.40$              & $23.17$              & $62.94$      & $60.46$     \\
\bottomrule
\end{tabular}
\caption{Model Performance on \textsc{OpenRewriteEval} formality category.}
\label{tab:openrewriteeval_formality}
\end{table*}

\begin{table*}[h]
\centering
\scriptsize
% \fontsize{9pt}{9pt}\selectfont
% \setlength{\tabcolsep}{1.5pt}
\begin{tabular}{lrrrrrrrrr}
\toprule
$D_\text{Paraphrase}$ &
  \multicolumn{1}{l}{} &
  \multicolumn{1}{l}{} &
  \multicolumn{2}{c}{NLI} &
  \multicolumn{1}{l}{} &
  \multicolumn{1}{l}{} &
  \multicolumn{1}{l}{} &
  \multicolumn{2}{c}{ROUGE-L} \\\cmidrule{4-5} \cmidrule{9-10}
                                & Edit Ratio & Len Ratio & s-p    & p-s    & SARI    & BLEU    & GLEU   & All     & Updated \\ \hline
                                \multicolumn{10}{l}{\cellcolor[HTML]{DDDDDD}Pretrained LLMs}      \\
PaLM-8B                         & $0.21$    & $0.35$    & $0.30$ & $0.12$ & $25.86$ & $1.29$  & $0.34$ & $5.85$  & $4.66$  \\
PaLM-62B                        & $0.27$    & $1.07$    & $0.28$ & $0.18$ & $28.18$ & $3.84$  & $0.31$ & $14.24$ & $11.09$ \\
PaLM 2-S                         & $0.73$    & $3.14$    & $0.49$ & $0.28$ & $28.34$ & $1.91$  & $0.19$ & $8.69$  & $8.02$  \\
LLaMA-65B                       & $0.84$    & $4.87$    & $0.84$ & $0.83$ & $27.19$ & $9.88$  & $1.29$ & $23.46$ & $17.92$ \\ \hline
\multicolumn{10}{l}{\cellcolor[HTML]{DDDDDD}Instruction-Tuned}      \\
Alpaca-7B                       & $0.10$    & $0.77$    & $0.98$ & $0.93$ & $37.38$ & $18.76$ & $4.25$ & $41.28$ & $36.27$ \\
Alpaca-13B                      & $0.10$    & $0.83$    & $0.98$ & $0.95$ & $39.18$ & $21.82$ & $6.07$ & $44.74$ & $39.77$ \\
Alpaca-PaLM 2-S & $0.13$ & $0.78$ & $0.98$ & $0.93$ & $39.92$ & $14.73$ & $5.18$ & $38.71$ & $37.67$ \\
Vicuna-7B                       & $0.15$    & $0.89$    & $0.97$ & $0.92$ & $39.77$ & $13.38$ & $4.71$ & $34.81$ & $34.38$ \\
Vicuna-13B                      & $0.16$    & $0.99$    & $0.97$ & $0.91$ & $39.63$ & $13.15$ & $4.75$ & $35.12$ & $34.46$ \\ 
Flan-PaLM-62B                       & $0.07$    & $0.67$    & $0.98$ & $0.74$ & $25.32$ & $24.31$ & $3.09$ & $44.46$ & $6.12$  \\
%ChatGPT & $0.13$ & $0.94$ & $0.93$ & $0.9$ & $42.51$ & $16.98$ & $8.01$ & $39.19$ & $38.7$ \\
InsGPT & $0.11$ & $0.53$ & $0.56$ & $0.55$ & $25.93$ & $17.19$ & $1.38$ & $30.46$ & $15.26$ \\
\hline
\multicolumn{10}{l}{\cellcolor[HTML]{DDDDDD}RewriteLMs}      \\
Rewrite-PaLM-62B                    & $0.10$    & $1.02$    & $0.96$ & $0.90$ & $33.98$ & $23.99$ & $3.16$ & $46.95$ & $35.99$ \\
Rewrite-Flan-PaLM-62B               & $0.09$    & $0.90$    & $0.96$ & $0.87$ & $36.16$ & $24.56$ & $4.98$ & $47.31$ & $38.53$ \\
Rewrite-PaLM 2-S                 & $0.10$    & $1.00$    & $0.97$ & $0.93$ & $39.53$ & $23.87$ & $5.51$ & $47.04$ & $43.26$ \\ 
Rewrite-RL-PaLM 2-S & $0.11$    & $0.99$    & $0.98$ & $0.92$ & $40.29$ & $22.66$ & $5.36$ & $45.67$ & $42.70$ \\
Rewrite-RL$_\text{r/w}$-PaLM 2-S  & $0.17$    & $1.37$    & $0.98$ & $0.94$ & $40.55$ & $22.52$ & $5.35$ & $45.39$ & $41.62$ \\
\bottomrule
\end{tabular}
\caption{Model Performance on \textsc{OpenRewriteEval} paraphrase category.}
\label{tab:openrewriteeval_paraphrase}
\end{table*}

\begin{table*}[h]
\centering
\scriptsize
% \fontsize{9pt}{9pt}\selectfont
% \setlength{\tabcolsep}{1.5pt}
\begin{tabular}{lrrrrrrrrr}
\toprule
$D_\text{Shorten}$ &
  \multicolumn{1}{l}{} &
  \multicolumn{1}{l}{} &
  \multicolumn{2}{c}{NLI} &
  \multicolumn{1}{l}{} &
  \multicolumn{1}{l}{} &
  \multicolumn{1}{l}{} &
  \multicolumn{2}{c}{ROUGE-L} \\\cmidrule{4-5} \cmidrule{9-10}
                                & Edit Ratio & Len Ratio & s-p    & p-s    & SARI    & BLEU    & GLEU    & All     & Updated \\ \hline
                                \multicolumn{10}{l}{\cellcolor[HTML]{DDDDDD}Pretrained LLMs}      \\
PaLM-8B                         & $0.22$    & $0.34$    & $0.29$ & $0.08$ & $22.51$ & $1.21$  & $0.60$  & $5.14$  & $4.37$  \\
PaLM-62B                        & $0.32$    & $1.29$    & $0.28$ & $0.12$ & $26.28$ & $2.25$  & $0.87$  & $12.31$ & $11.13$ \\
PaLM 2-S                         & $1.12$    & $5.21$    & $0.63$ & $0.35$ & $26.21$ & $2.49$  & $0.33$  & $8.92$  & $6.92$  \\
LLaMA-65B                       & $0.76$    & $4.55$    & $0.85$ & $0.82$ & $27.87$ & $14.14$ & $3.51$  & $28.03$ & $21.55$ \\ \hline
\multicolumn{10}{l}{\cellcolor[HTML]{DDDDDD}Instruction-Tuned}      \\
Alpaca-7B                       & $0.12$    & $0.58$    & $0.97$ & $0.87$ & $36.41$ & $22.88$ & $8.47$  & $46.67$ & $42.15$ \\
Alpaca-13B                      & $0.12$    & $0.65$    & $0.97$ & $0.95$ & $37.38$ & $24.32$ & $11.14$ & $48.26$ & $43.42$ \\
Alpaca-PaLM 2-S & $0.15$ & $0.45$ & $0.95$ & $0.8$ & $34.26$ & $12.37$ & $8.5$ & $36.51$ & $35.53$ \\

Vicuna-7B                       & $0.18$    & $0.86$    & $0.94$ & $0.81$ & $34.48$ & $13.55$ & $7.47$  & $35.36$ & $34.10$ \\
Vicuna-13B                      & $0.16$    & $0.77$    & $0.97$ & $0.87$ & $35.70$ & $16.51$ & $8.94$  & $39.41$ & $37.57$ \\
Flan-PaLM-62B                       & $0.09$    & $0.57$    & $0.93$ & $0.59$ & $25.98$ & $28.72$ & $4.84$  & $48.27$ & $5.67$  \\ 
%ChatGPT & $0.13$ & $0.68$ & $0.98$ & $0.93$ & $38.67$ & $21.2$ & $13.49$ & $45.49$ & $44.26$ \\
InsGPT & $0.09$ & $0.6$ & $0.65$ & $0.62$ & $27.45$ & $28.81$ & $5.91$ & $45.28$ & $24.67$ \\
\hline
\multicolumn{10}{l}{\cellcolor[HTML]{DDDDDD}RewriteLMs}      \\
Rewrite-PaLM-62B                    & $0.10$    & $0.73$    & $0.97$ & $0.85$ & $37.46$ & $32.03$ & $11.75$ & $54.97$ & $44.30$ \\
Rewrite-Flan-PaLM-62B               & $0.11$    & $0.60$    & $0.95$ & $0.79$ & $38.09$ & $27.61$ & $11.55$ & $51.49$ & $42.30$ \\
Rewrite-PaLM 2-S                 & $0.12$    & $0.65$    & $0.97$ & $0.82$ & $38.55$ & $27.11$ & $10.61$ & $51.75$ & $44.84$ \\ 
Rewrite-RL-PaLM 2-S & $0.12$    & $0.69$    & $0.98$ & $0.84$ & $38.40$ & $26.92$ & $10.39$ & $51.39$ & $44.64$ \\
Rewrite-RL$_\text{r/w}$-PaLM 2-S  & $0.16$    & $0.94$    & $1.00$ & $0.92$ & $39.50$ & $28.99$ & $11.84$ & $53.11$ & $46.75$ \\
\bottomrule
\end{tabular}
\caption{Model Performance on \textsc{OpenRewriteEval} shorten category.}
\label{tab:openrewriteeval_shorten}
\end{table*}

\begin{table*}[h]
\centering
\scriptsize
% \fontsize{9pt}{9pt}\selectfont
% \setlength{\tabcolsep}{1.5pt}
\begin{tabular}{lrrrrrrrrr}
\toprule
$D_\text{Elaborate}$ &
  \multicolumn{1}{l}{} &
  \multicolumn{1}{l}{} &
  \multicolumn{2}{c}{NLI} &
  \multicolumn{1}{l}{} &
  \multicolumn{1}{l}{} &
  \multicolumn{1}{l}{} &
  \multicolumn{2}{c}{ROUGE-L} \\\cmidrule{4-5} \cmidrule{9-10}
                                & Edit Ratio & Len Ratio & s-p    & p-s    & SARI    & BLEU    & GLEU   & All     & Updated \\ \hline
                                \multicolumn{10}{l}{\cellcolor[HTML]{DDDDDD}Pretrained LLMs}      \\
PaLM-8B                         & $0.21$    & $0.33$    & $0.30$ & $0.15$ & $20.88$ & $0.85$  & $0.30$ & $6.16$  & $3.79$  \\
PaLM-62B                        & $0.29$    & $1.03$    & $0.33$ & $0.08$ & $23.32$ & $1.23$  & $0.38$ & $10.61$ & $9.25$  \\
PaLM 2-S                         & $1.24$    & $5.85$    & $0.72$ & $0.40$ & $26.58$ & $2.32$  & $1.09$ & $10.79$ & $9.46$  \\
LLaMA-65B                       & $0.61$    & $3.78$    & $0.83$ & $0.86$ & $28.80$ & $11.56$ & $3.90$ & $29.51$ & $17.94$ \\ \hline
\multicolumn{10}{l}{\cellcolor[HTML]{DDDDDD}Instruction-Tuned}      \\
Alpaca-7B                       & $0.18$    & $1.04$    & $0.73$ & $0.57$ & $30.63$ & $6.19$  & $3.06$ & $23.74$ & $18.63$ \\
Alpaca-13B                      & $0.18$    & $1.09$    & $0.72$ & $0.62$ & $31.67$ & $7.81$  & $4.72$ & $26.01$ & $18.73$ \\
Alpaca-PaLM 2-S & $0.17$ & $0.85$ & $0.6$ & $0.49$ & $30.28$ & $5.32$ & $3.2$ & $20.98$ & $16.95$ \\

Vicuna-7B                       & $0.46$    & $2.73$    & $0.88$ & $0.56$ & $31.74$ & $5.01$  & $2.53$ & $24.21$ & $18.41$ \\
Vicuna-13B                      & $0.46$    & $2.69$    & $0.89$ & $0.50$ & $31.71$ & $4.80$  & $2.57$ & $24.03$ & $18.20$ \\
Flan-PaLM-62B                       & $0.16$    & $0.24$    & $0.73$ & $0.26$ & $23.00$ & $2.31$  & $0.54$ & $13.42$ & $3.29$  \\
%ChatGPT & $0.17$ & $0.96$ & $0.94$ & $0.75$ & $31.43$ & $5$ & $2.77$ & $23.04$ & $21.14$ \\
InsGPT & $0.16$ & $0.33$ & $0.32$ & $0.34$ & $22.19$ & $4.23$ & $1.6$ & $13.6$ & $5.55$ \\
\hline
\multicolumn{10}{l}{\cellcolor[HTML]{DDDDDD}RewriteLMs}      \\
Rewrite-PaLM-62B                    & $0.36$    & $2.02$    & $0.67$ & $0.38$ & $29.43$ & $6.30$  & $3.06$ & $26.63$ & $16.44$ \\
Rewrite-Flan-PaLM-62B               & $0.36$    & $2.04$    & $0.68$ & $0.35$ & $29.01$ & $5.07$  & $1.84$ & $24.77$ & $17.92$ \\
Rewrite-PaLM 2-S                 & $0.70$    & $3.84$    & $0.93$ & $0.53$ & $31.55$ & $5.66$  & $3.11$ & $26.17$ & $17.58$ \\ 
Rewrite-RL-PaLM 2-S & $0.79$    & $4.23$    & $0.97$ & $0.55$ & $32.39$ & $5.83$  & $3.11$ & $26.32$ & $18.13$ \\
Rewrite-RL$_\text{r/w}$-PaLM 2-S  & $0.74$    & $4.22$    & $0.99$ & $0.77$ & $33.25$ & $8.67$  & $3.75$ & $30.05$ & $20.15$ \\
\bottomrule
\end{tabular}
\caption{Model Performance on \textsc{OpenRewriteEval} elaborate category.}
\label{tab:openrewriteeval_elaborate}
\end{table*}

\begin{table*}[h]
\centering
\scriptsize
% \fontsize{9pt}{9pt}\selectfont
% \setlength{\tabcolsep}{1.5pt}
\begin{tabular}{lrrrrrrrrr}
\toprule
$D_\text{MixedWiki}$ &
  \multicolumn{1}{l}{} &
  \multicolumn{1}{l}{} &
  \multicolumn{2}{c}{NLI} &
  \multicolumn{1}{l}{} &
  \multicolumn{1}{l}{} &
  \multicolumn{1}{l}{} &
  \multicolumn{2}{c}{ROUGE-L} \\\cmidrule{4-5} \cmidrule{9-10}
                                & Edit Ratio & Len Ratio & s-p    & p-s    & SARI    & BLEU    & GLEU    & All     & Updated \\ \hline
                                \multicolumn{10}{l}{\cellcolor[HTML]{DDDDDD}Pretrained LLMs}      \\
PaLM-8B                         & $0.33$    & $1.63$    & $0.31$ & $0.12$ & $28.00$ & $3.47$  & $1.15$  & $14.41$ & $13.48$ \\
PaLM-62B                        & $0.33$    & $1.56$    & $0.21$ & $0.11$ & $27.55$ & $3.39$  & $1.10$  & $13.60$ & $12.81$ \\
PaLM 2-S                         & $1.41$    & $7.08$    & $0.67$ & $0.41$ & $28.21$ & $2.39$  & $0.56$  & $8.47$  & $8.83$  \\
LLaMA-65B                       & $0.68$    & $4.18$    & $0.84$ & $0.87$ & $29.04$ & $14.44$ & $2.69$  & $27.54$ & $24.44$ \\ \hline
\multicolumn{10}{l}{\cellcolor[HTML]{DDDDDD}Instruction-Tuned}      \\
Alpaca-7B                       & $0.09$    & $0.94$    & $0.93$ & $0.91$ & $35.73$ & $31.86$ & $8.76$  & $50.17$ & $38.02$ \\
Alpaca-13B                      & $0.08$    & $0.95$    & $0.92$ & $0.91$ & $35.87$ & $32.45$ & $9.00$  & $50.49$ & $38.41$ \\
Alpaca-PaLM2-S & $0.1$ & $0.91$ & $0.94$ & $0.92$ & $39.96$ & $31.75$ & $12.75$ & $50.24$ & $42.8$ \\

Vicuna-7B                       & $0.20$    & $1.43$    & $0.90$ & $0.86$ & $39.34$ & $23.11$ & $9.51$  & $41.00$ & $38.09$ \\
Vicuna-13B                      & $0.21$    & $1.50$    & $0.92$ & $0.86$ & $39.75$ & $23.72$ & $10.03$ & $41.79$ & $39.06$ \\
Flan-PaLM-62B                       & $0.20$    & $0.83$    & $0.11$ & $0.09$ & $24.73$ & $4.57$  & $0.72$  & $14.26$ & $9.20$  \\ 
%ChatGPT & $0.11$ & $0.95$ & $0.97$ & $0.92$ & $40.69$ & $26.28$ & $11.86$ & $45.9$ & $43.74$ \\
InsGPT & $0.07$ & $0.71$ & $0.76$ & $0.7$ & $28.76$ & $30.07$ & $5.64$ & $44.53$ & $23.14$ \\\hline
\multicolumn{10}{l}{\cellcolor[HTML]{DDDDDD}RewriteLMs}      \\
Rewrite-PaLM-62B                    & $0.09$    & $0.98$    & $0.95$ & $0.84$ & $38.54$ & $35.43$ & $10.20$ & $53.95$ & $42.27$ \\
Rewrite-Flan-PaLM-62B               & $0.10$    & $0.93$    & $0.93$ & $0.78$ & $39.71$ & $33.89$ & $10.71$ & $52.65$ & $42.88$ \\
Rewrite-PaLM 2-S                 & $0.13$    & $1.05$    & $0.93$ & $0.83$ & $42.81$ & $32.52$ & $13.35$ & $51.28$ & $46.47$ \\
Rewrite-RL-PaLM 2-S & $0.14$    & $1.08$    & $0.93$ & $0.84$ & $42.50$ & $32.41$ & $12.80$ & $50.79$ & $46.38$ \\
Rewrite-RL$_\text{r/w}$-PaLM 2-S  & $0.15$    & $1.21$    & $0.94$ & $0.87$ & $42.93$ & $34.42$ & $13.48$ & $52.25$ & $47.12$ \\
\bottomrule
\end{tabular}
\caption{Model Performance on \textsc{OpenRewriteEval} mixed Wiki category.}
\label{tab:openrewriteeval_wiki}
\end{table*}

\begin{table*}[!t]
\centering
\scriptsize
% \fontsize{9pt}{9pt}\selectfont
% \setlength{\tabcolsep}{1.5pt}
\begin{tabular}{lrrrrrrrrr}
\toprule
$D_\text{MixedOthers}$ &
  \multicolumn{1}{l}{} &
  \multicolumn{1}{l}{} &
  \multicolumn{2}{c}{NLI} &
  \multicolumn{1}{l}{} &
  \multicolumn{1}{l}{} &
  \multicolumn{1}{l}{} &
  \multicolumn{2}{c}{ROUGE-L} \\ \cmidrule{4-5} \cmidrule{9-10}
                                & Edit Ratio & Len Ratio & s-p    & p-s    & SARI    & BLEU    & GLEU   & All     & Updated \\ \hline
                                \multicolumn{10}{l}{\cellcolor[HTML]{DDDDDD}Pretrained LLMs}      \\
PaLM-8B                         & $0.22$    & $0.57$    & $0.31$ & $0.12$ & $26.73$ & $1.96$  & $0.20$ & $7.31$  & $6.15$  \\
PaLM-62B                        & $0.27$    & $1.10$    & $0.28$ & $0.10$ & $30.71$ & $2.28$  & $0.41$ & $13.34$ & $11.86$ \\
PaLM 2-S                         & $0.97$    & $4.47$    & $0.60$ & $0.30$ & $30.49$ & $1.25$  & $0.23$ & $7.62$  & $7.48$  \\
LLaMA-65B                       & $0.56$    & $3.54$    & $0.79$ & $0.76$ & $26.85$ & $8.86$  & $0.96$ & $23.83$ & $19.40$ \\ \hline
\multicolumn{10}{l}{\cellcolor[HTML]{DDDDDD}Instruction-Tuned}      \\
Alpaca-7B                       & $0.11$    & $0.88$    & $0.85$ & $0.75$ & $34.03$ & $15.82$ & $1.78$ & $37.27$ & $28.52$ \\
Alpaca-13B                      & $0.11$    & $0.88$    & $0.86$ & $0.74$ & $33.42$ & $16.53$ & $2.29$ & $38.00$ & $27.24$ \\
Alpaca-PaLM 2-S & $0.13$ & $0.82$ & $0.85$ & $0.76$ & $36.89$ & $13.79$ & $3.1$ & $34.5$ & $29.51$ \\
Vicuna-7B                       & $0.23$    & $1.30$    & $0.75$ & $0.56$ & $38.51$ & $10.20$ & $3.07$ & $28.47$ & $26.35$ \\
Vicuna-13B                      & $0.23$    & $1.36$    & $0.78$ & $0.59$ & $39.05$ & $11.35$ & $3.21$ & $30.11$ & $27.42$ \\
Flan-PaLM-62B                       & $0.09$    & $0.57$    & $0.83$ & $0.59$ & $24.58$ & $14.37$ & $0.94$ & $32.83$ & $3.60$  \\ 
%ChatGPT & $0.15$ & $0.86$ & $0.89$ & $0.76$ & $39.87$ & $12.46$ & $4.09$ & $32.02$ & $30.71$ \\
InsGPT & $0.11$ & $0.48$ & $0.54$ & $0.48$ & $25.53$ & $12.12$ & $0.78$ & $25.31$ & $11.6$ \\
\hline
\multicolumn{10}{l}{\cellcolor[HTML]{DDDDDD}RewriteLMs}      \\
Rewrite-PaLM-62B                    & $0.21$    & $1.47$    & $0.76$ & $0.63$ & $34.25$ & $13.99$ & $2.16$ & $34.89$ & $25.63$ \\
Rewrite-Flan-PaLM-62B               & $0.22$    & $1.46$    & $0.78$ & $0.59$ & $34.35$ & $12.40$ & $1.90$ & $32.54$ & $25.44$ \\
Rewrite-PaLM 2-S                 & $0.41$    & $2.36$    & $0.88$ & $0.68$ & $36.85$ & $10.30$ & $2.00$ & $29.82$ & $25.42$ \\ 
Rewrite-RL-PaLM 2-S & $0.45$    & $2.59$    & $0.89$ & $0.71$ & $36.85$ & $9.66$  & $1.85$ & $29.26$ & $25.55$ \\
Rewrite-RL$_\text{r/w}$-PaLM 2-S  & $0.50$    & $2.92$    & $0.94$ & $0.83$ & $35.16$ & $10.28$ & $1.47$ & $30.36$ & $26.32$ \\
\bottomrule
\end{tabular}
\caption{Model Performance on \textsc{OpenRewriteEval} mixed others category.}
\label{tab:openrewriteeval_others}
\end{table*}

\bibliographystyle{iclr2024_conference}

\end{document}